\documentclass{article}

\usepackage[preprint]{neurips_2024}

\usepackage[utf8]{inputenc} 
\usepackage[T1]{fontenc}    
\usepackage{hyperref}       
\usepackage{url}            
\usepackage{booktabs}       
\usepackage{amsfonts}       
\usepackage{nicefrac}       
\usepackage{microtype}      
\usepackage{xcolor}         
\usepackage{bbding}         
\usepackage{multirow}       
\usepackage{longtable}      
\usepackage{caption}        
\usepackage{graphicx}       
\usepackage{subfigure}      
\usepackage{float}          
\usepackage{placeins}       
\usepackage{makecell}       
\usepackage{bm}             
\usepackage{CJKutf8}        
\usepackage{threeparttable} 

\title{
\begin{minipage}{0.1\textwidth}
\includegraphics[width=0.8\linewidth]{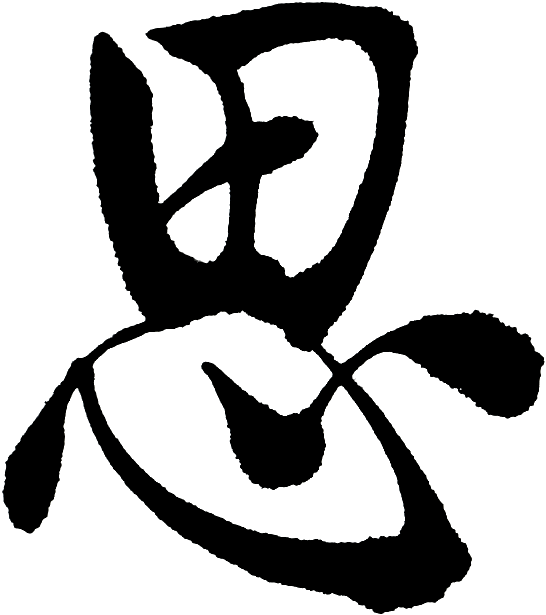}
\end{minipage}
\begin{minipage}{0.75\textwidth}
SI-SD: Sleep Interpreter through awake-guided cross-subject Semantic Decoding
\end{minipage}
}

\author{Hui Zheng\textsuperscript{*,2,4}, Zhong-Tao Chen\textsuperscript{*,1}, Hai-Teng Wang\textsuperscript{1},\\
\textbf{Jian-Yang Zhou\textsuperscript{3,4}, Lin Zheng\textsuperscript{1}, Pei-Yang Lin\textsuperscript{1}, Yun-Zhe Liu\textsuperscript{\dag,1,4}}\\
\textsuperscript{1}Beijing Normal University, \textsuperscript{2}Peking University,\\
\textsuperscript{3}Capital Medical University, \textsuperscript{4}Chinese Institute for Brain Research\\
\texttt{*Equal contribution,\dag yunzhe.liu@bnu.edu.cn}
}

\begin{document}

\maketitle

\begin{abstract}
  Understanding semantic content from brain activity during sleep represents a major goal in neuroscience. While studies in rodents have shown spontaneous neural reactivation of memories during sleep, capturing the semantic content of human sleep poses a significant challenge due to the absence of well-annotated sleep datasets and the substantial differences in neural patterns between wakefulness and sleep. To address these challenges, we designed a novel cognitive neuroscience experiment and collected a comprehensive, well-annotated electroencephalography (EEG) dataset from 134 subjects during both wakefulness and sleep. Leveraging this benchmark dataset, we developed SI-SD\footnote{SI refers to the phonetic transcription of "\begin{CJK}{UTF8}{bsmi}思\end{CJK}" (i.e., think) in Chinese.} that enhances sleep semantic decoding through the position-wise alignment of neural latent sequence between wakefulness and sleep. In the 15-way classification task, our model achieves 24.12\% and 21.39\% top-1 accuracy on unseen subjects for NREM 2/3 and REM sleep, respectively, surpassing all other baselines. With additional fine-tuning, decoding performance improves to 30.32\% and 31.65\%, respectively. Besides, inspired by previous neuroscientific findings, we systematically analyze how the "Slow Oscillation" event impacts decoding performance in NREM 2/3 sleep -- decoding performance on unseen subjects further improves to 40.02\%. Together, our findings and methodologies contribute to a promising neuro-AI framework for decoding brain activity during sleep.
\end{abstract}

\section{Introduction}
Sleep plays a fundamental role in memory consolidation \cite{klinzing2019mechanisms,brodt2023sleep}. Past memories are known to reactivate during sleep, especially during the N2/3 stage of non-rapid eye-movement (NREM) sleep \cite{ngo2022shaping}. In rodents, hippocampal cells have been found to replay their firing patterns during sleep, recapitulating awake experiences in a time-compressed order \cite{wilson1994reactivation,skaggs1996replay}. In humans, while direct cell recordings are rare, recording scalp electroencephalography (EEG) during sleep is possible. Recent work on human sleep decoding has identified endogenous memory reactivation during the N2/3 stage of sleep, the extent of which was positively related to subsequent memory performance \cite{schreiner2021endogenous}.

Despite the significance of sleep decoding in humans, attempts of this sort are scarce in both the neuroscience and computer science communities. This is because there is no well-annotated sleep dataset that provides clear ground-truth information about which memory is activated and when during sleep. The neural activity during wakefulness and sleep differ greatly, causing classifiers trained on awake periods to struggle when applied to sleep periods \cite{liu2022decoding}. Generalizing neural representation across subjects is especially challenging during sleep due to the spontaneous nature of memory reactivation without timed neural responses.

\begin{figure}[h]
  \centering
  \includegraphics[width=\linewidth]{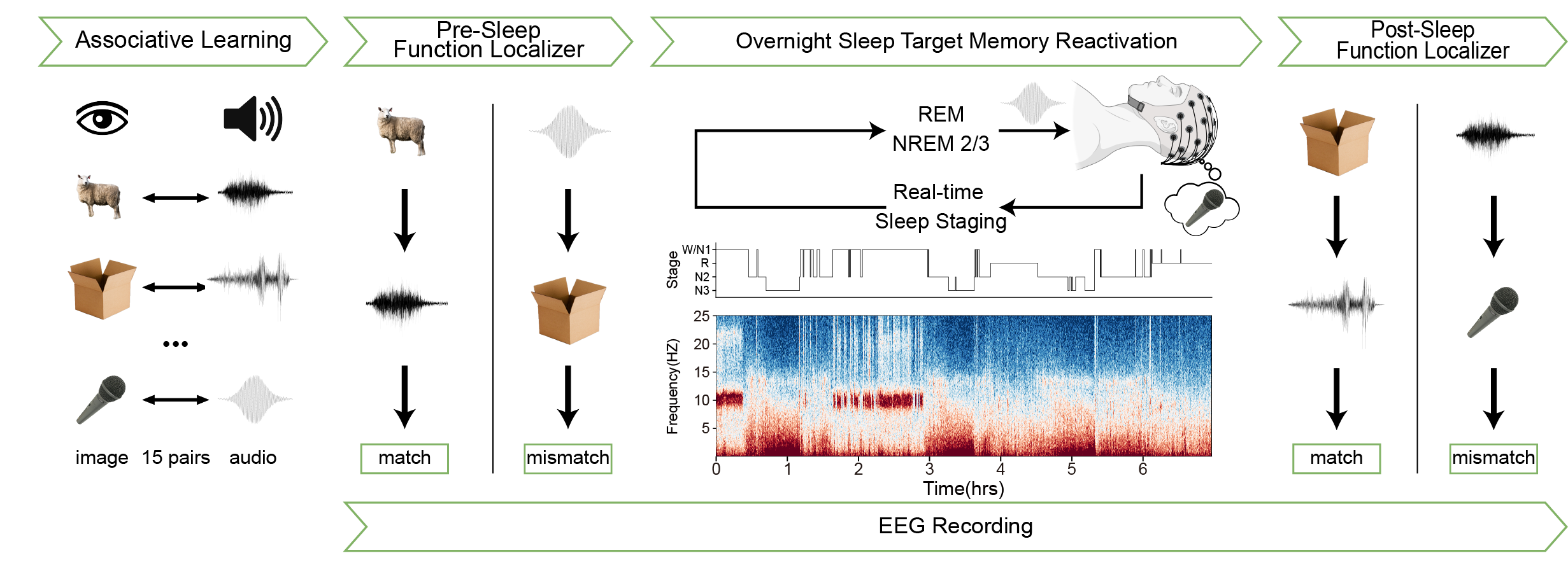}
  \caption{The experiment design for sleep semantic decoding. During the experiment, each subject must memorize the shared set of semantic concepts (including 15 pre-determined semantic concepts, each paired with both an image stimulus and an audio stimulus); see Appendix \ref{sec:supp-expr-design} for more details.}
  \label{fig:expr-design}
\end{figure}

To capture the content of neural reactivation in humans, we designed an innovative cognitive neuroscience experiment based on the classical targeted memory reactivation (TMR) paradigm \cite{rasch2007odor}, as illustrated in Figure \ref{fig:expr-design}. We employed a close-loop stimulation system allowing for real-time, automatic sleep staging \cite{vallat2021open}. When subjects reached NREM 2/3 and REM sleep stages, auditory stimuli from 15 pre-determined semantic concepts were played every 4-6 seconds, with concurrent whole-brain EEG recordings. This approach provided precise timing and content of memory reactivation during sleep, facilitating the training of a neural decoder on these cued sleep intervals. Before and after sleep, subjects were asked to recall the audio-image pairs, enabling the position-wise alignment of neural latent sequence elicited by the same stimuli during both wakefulness and sleep.

Based on this dataset, we introduce SI-SD, a framework for decoding semantic information in NREM 2/3 and REM sleep stages. SI-SD learns subject-agnostic representations in a supervised manner across subjects, offering off-the-shelf sleep semantic decoding capability for unseen subjects. Given the challenges of sleep data collection, we align the semantic process during wakefulness and sleep to enhance sleep semantic decoding. Ablation studies show that the neural representation sequence from awake signals provides a better template to extract semantic components from the noisy EEG signals during sleep, compared to that from other modalities (e.g., CLIP \cite{radford2021learning} or wav2vec 2.0 \cite{baevski2020wav2vec}). In the 15-way classification task, our model achieves 24.12\% and 21.39\% decoding performance on unseen subjects for NREM 2/3 and REM sleep, surpassing its baseline by 3.11\% and 4.67\%, respectively. With additional data for fine-tuning, our model achieves 30.32\% and 31.65\%, respectively.

Besides, inspired by previous neuroscientific findings, we examine the impact of the "Slow Oscillation" event on decoding performance during NREM 2/3 sleep. Our results validate existing neuroscience findings \cite{schreiner2021endogenous} while also uncovering new features. Notably, in a sub-dataset where auditory stimuli are presented on the transition to the first "Up" of the "Slow Oscillation" event, decoding performance reaches up to 40.02\% for unseen subjects. The transfer experiments from wakefulness to sleep further prove the presence of shared semantic components between them. It marks a promising neuro-inspired AI approach \cite{saxe2021if,richards2019deep} for neural decoding.

To sum up, the main contributions of our work comprise:
\begin{enumerate}
  \item A well-annotated whole-brain EEG dataset over 134 human subjects, containing both awake and overnight sleep data with ground truth regarding which memory was activated and when. The dataset will be publicly available.
  \item A novel framework for EEG sleep semantic decoding task -- SI-SD, aligning neural latent sequence position-wise between wakefulness and sleep to enhance sleep semantic decoding.
  \item Demonstration that SI-SD offers an off-the-shelf, subject-agnostic sleep semantic decoding model -- SI-SD achieves up to 40.02\% and 21.39\% top-1 accuracy on unseen subjects for NREM 2/3 and REM sleep, surpassing all other baselines.
  \item Comprehensive analysis of how the "Slow Oscillation" event impacts decoding performance -- validating existing neuroscience findings while also uncovering new features.
\end{enumerate}

\section{Related Works}
\subsection{Neural Basis of Sleep Deoding}
During sleep, humans are largely unconscious, and neural reactivation occurs spontaneously \cite{siclari2017neural,schonauer2017decoding}. Consequently, gathering a well-annotated dataset that offers precise timing and content of neural reactivation during sleep is challenging, posing a significant hurdle in sleep decoding research. To address this challenge, current studies can mainly be categorized into two streams, depending on whether or not they give subjects external stimulation.

Some studies \cite{horikawa2013neural,siclari2017neural} extract memory reactivation content by soliciting reports from subjects after awakening them. However, the amount of the obtained dataset is far less than what is required to train the model, resulting in the sleep decoder being trained on data from wakefulness \cite{horikawa2013neural}. Besides, this approach is limited solely to the REM sleep stage, where neural patterns resemble wakefulness. Considering the neural patterns during the NREM sleep stage exhibit even greater differences from wakefulness, this approach is unsuitable for decoding memory content from the NREM sleep stage.

Other studies \cite{turker2022behavioral} give subjects external stimuli to reactivate targeted memory, which is commonly referred to as Target Memory Reactivation (TMR). Sleep has classically been considered a time when we are disconnected from the world, with significantly reduced (or absent) reactivity to external stimuli. However, several studies in recent years indicate that sleepers can process external stimuli at different cognitive levels, encompassing semantic and decision-making stages, rather than merely the low-level sensory processing \cite{strauss2015disruption,issa2011altered,kouider2014inducing}. Therefore, we follow these studies to design our cognitive neuroscience experiment; see Appendix \ref{sec:supp-expr-design} for more details.

\subsection{Contrastive Learning in BCI}
Recently, the field of representation learning has grown rapidly with the powerful contrastive learning \cite{chen2020simple}. Khosla et al. \cite{khosla2020supervised} further extend it to the supervised setting, thus allowing the model to learn more discriminative representations. Along with the recent development of large models \cite{baevski2020wav2vec,radford2021learning,lewis2019bart}, many studies utilize the similarity of representations between the human brain and large models to improve the performance of neural decoding \cite{defossez2023decoding,du2023decoding,duan2023dewave}. These studies are primarily categorized into two groups based on the representation units utilized to calculate contrastive loss.

Some studies \cite{du2023decoding,song2023decoding} treat the entire sequence of neural representations as the basic unit for contrastive learning. Both BraVL\cite{du2023decoding} and NICE \cite{song2023decoding} project the encoded neural representation sequence into a representation unit and enhance decoding performance by aligning the representation unit with that of CLIP \cite{radford2021learning}. However, this approach often demands numerous semantic categories, making it impractical for sleep decoding experiments given the constraints of human cognitive load.

Other studies \cite{defossez2023decoding,duan2023dewave,cho2023neural} decompose the entire neural representation sequence into neural representation units and perform contrastive learning at a more fine-grained level. D$\mathrm{\acute{e}}$fossez et al. \cite{defossez2023decoding} directly aligns the neural representation sequence with that of wav2vec 2.0 \cite{baevski2020wav2vec}, which greatly enhances the performance of speech perception decoding. DeWave \cite{du2023decoding} aligns with BART \cite{lewis2019bart} to decode language. NLA \cite{cho2023neural} adopts a distinct approach -- aligning the neural representation sequences across trials for the same category to improve speech decoding.

\section{Method}
\subsection{Task Definition}
Following our designed cognitive experiment paradigm illustrated in Figure \ref{fig:expr-design}, we record three different kinds (i.e., image-evoked, audio-evoked, and TMR-related\footnote{We refer to audio-evoked EEG signals during sleep as TMR-related, instead of TMR-evoked, as our experiment differs from the TMR paradigm. In the traditional TMR paradigm, subjects are presented with selected auditory stimuli, instead of all auditory stimuli learned during the associative learning stage.}) of EEG signals from each subject. Despite originating from different types of stimulation, all these EEG signals capture the semantic processing of the shared semantic concepts within the brain.

We formulate the EEG signals as $\mathcal{X}\in\mathbb{R}^{C\times T}$, where $C$ is the number of EEG channels and $T$ is the total timestamps. The associated semantic label is denoted as $\bm{y}\in\mathcal{Y}$, where $\mathcal{Y}$ represents the set of 15 pre-determined semantic concepts (each paired with both an audio stimulus and an image stimulus). In summary, for each subject, the total dataset contains three sub-datasets, i.e., $\mathcal{D}_{s}=\langle \mathcal{D}^{img}_{s},\mathcal{D}^{aud}_{s},\mathcal{D}^{tmr}_{s}\rangle,s\in\mathcal{S}$, where $\mathcal{S}$ is the set of subjects and each sub-dataset comprises paired EEG-concept data ($\langle \mathcal{X},\bm{y}\rangle$). Therefore, the goal of this work is to decode the corresponding semantic label $\bm{y}$ from TMR-related raw EEG signals $\mathcal{X}^{tmr}$, believing that the incorporation of datasets from various subjects, coupled with awake signals $\langle \mathcal{X}^{img},\mathcal{X}^{aud}\rangle$ together is beneficial for learning generalizable and discriminative representations for sleep semantic decoding.

\subsection{Model Architecture}
\begin{figure}[h]
  \centering
  \includegraphics[width=\linewidth]{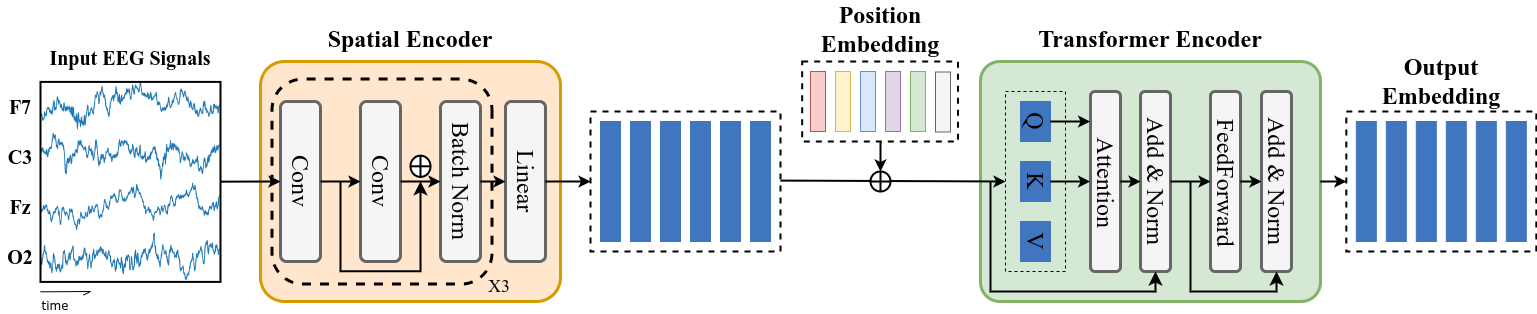}
  \caption{The overall architecture of Neural Transformer.}
  \label{fig:neural-xfmr}
\end{figure}

We introduce the neural Transformer, a general architecture for EEG decoding tasks (i.e., decoding semantic label $\bm{y}$ from different kinds of EEG signals $\langle \mathcal{X}^{img},\mathcal{X}^{aud},\mathcal{X}^{tmr}\rangle$) in this work, as illustrated in Figure \ref{fig:neural-xfmr}. The model contains two parts: (1) Spatial Encoder, and (2) Transformer Encoder.

\paragraph{Spatial Encoder.}As each EEG sample $\mathcal{X}$ has multiple channels, it is vital to fuse multiple channels to extract meaningful tokens before token-wise interaction by self-attention. Our spatial encoder, comprising residual convolution blocks and a linear projection, encodes raw EEG signals into token embeddings. The residual convolution block contains two 1-D convolution layers and a batch normalization layer \cite{ioffe2015batch}. We denote the output token embeddings from the spatial encoder as
\begin{equation}
  \label{equ:spatial-encoder}
  \mathcal{E}_{t}=\{\bm{e}^{t}_{i}\in\mathbb{R}^{d}|i=1,...,N\},
\end{equation}
where $d$ is the dimension of the embeddings, and $N$ is the number of token embeddings.

\paragraph{Transformer Encoder.}Before feeding the embedding sequence into the transformer encoder, we add the parameter-free position embeddings introduced in \cite{vaswani2017attention} to enable the model to be aware of the relative positions among embeddings. Then, the sequence of embeddings will be fed into the transformer encoder to get the final encoded $\mathcal{E}=\{\bm{e}_{i}\in\mathbb{R}^{d}|i=1,...,N\}$. For the downstream 15-way classification task, we flatten the output embeddings followed by a classification head.

\subsection{Awake-Guided Latent Alignment}
Previous studies \cite{schreiner2021endogenous} on sleep semantic decoding show variations in decoding performance across stages of the "Slow Oscillation" event. We hypothesize that external stimuli evoke the semantic components with varying degrees of intensity during sleep, contributing to the presence of noisy EEG signals in the sleep dataset; see Section \ref{sec:so-analysis} for more details. Thus, relying solely on the sleep dataset for training may result in overfitting to noisy signals, compromising the robustness of representations.

In comparison, the awake dataset is more resource-rich, as we can obtain more samples with reliable annotations easily. Besides, the neural sequence induced by external stimuli during wakefulness demonstrates how the brain processes those stimuli, transitioning from low-level to high-level information. Therefore, the neural sequence serves as an ideal template for sleep semantic decoding.

\begin{figure}[h]
  \centering
  \includegraphics[width=\linewidth]{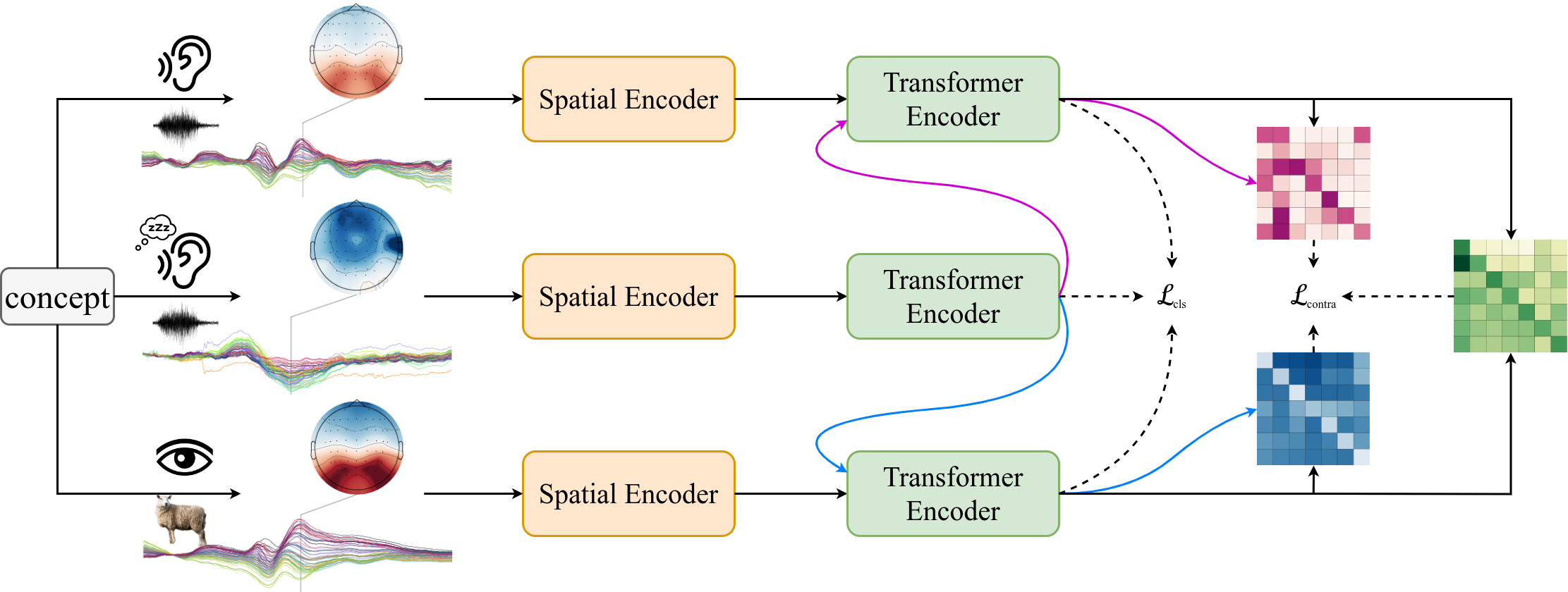}
  \caption{The overall architecture of SI-SD.}
  \label{fig:sisd}
\end{figure}

Since the neural patterns differ greatly among different tasks (as illustrated in Figure \ref{fig:sisd}), we utilize three different spatial encoders to extract meaningful tokens from raw EEG signals $\langle \mathcal{X}^{img},\mathcal{X}^{aud},\mathcal{X}^{tmr}\rangle$. Besides, since the intensity of semantic components during sleep is lower than wakefulness, sharing spatial encoders might mislead the tokenization of EEG signals during sleep. After tokenization, the embeddings are fed into the transformer encoder for sequence modeling. Then, the output embeddings $\mathcal{E}$ are used to predict the semantic label $\hat{\bm{y}}$ and calculate the classification loss.

To better utilize the template from the awake dataset, we introduce two special model designs: (1) a contrastive objective to align awake and sleep neural latent sequence position-wise, and (2) a shared encoder to further improve alignment. The overall architecture of SI-SD is illustrated in Figure \ref{fig:sisd}.

\paragraph{Contrastive Objective.}\textbf{Like NLA \cite{cho2023neural}, we calculate contrastive loss by decomposing the entire neural representation sequence into neural representation units, which perform contrastive learning at a more fine-grained level than NICE \cite{song2023decoding}.} The only difference is that we align the neural representation sequences sharing the same category, originating from different tasks. As illustrated in Figure \ref{fig:sisd}, our contrastive loss consists of three parts, each aligning representations from two different tasks.

For each batch $\mathcal{B}$, we randomly draw $|\mathcal{B}|$ sample pairs $\langle \mathcal{X}^{tmr},\bm{y}\rangle$ from the whole TMR-related dataset $\mathcal{D}^{tmr}=\{\mathcal{D}^{tmr}_{s}\}_{s\in\mathcal{S}}$. Then, for each sample pair, we further draw one image-evoked sample $\mathcal{X}^{img}$ and one audio-evoked sample $\mathcal{X}^{aud}$ according to the semantic label $\bm{y}$ and the subject id $s$. Thus, we have $\mathcal{B}=\{\langle \mathcal{X}^{img}_{i},\mathcal{X}^{aud}_{i},\mathcal{X}^{tmr}_{i},\bm{y}_{i}\rangle\}_{i=1}^{|\mathcal{B}|}$, where different items could come from different subjects. After mapping the raw EEG signals $\mathcal{X}$ into encoded embedding sequence $\mathcal{E}$, we utilize the \textbf{position-wise projection} to transform embeddings and \textbf{flatten} them to form the latent embedding $\bm{z}$. Then, we calculate the contrastive loss of three combinations according to the following equation:
\begin{equation}
  \mathcal{L}_{contra}=-\sum_{i\in\{1,...,|\mathcal{B}|\}}\frac{1}{|\mathcal{A}(i)|}\sum_{k\in\mathcal{A}(i)}\mathrm{log}\frac{\mathrm{e}^{\langle \bm{z}_{i},\bm{h}_{k}\rangle}}{\sum_{j\in\{1,...,|\mathcal{B}|\}}\mathrm{e}^{\langle \bm{z}_{i},\bm{h}_{j}\rangle}},
\end{equation}
where $\mathcal{A}(i)=\{k|k\in\{1,...,|\mathcal{B}|\},\bm{y}_{k}=\bm{y}_{i}\}$, $\langle \cdot,\cdot\rangle$ the inner product, and $\bm{z}$, $\bm{h}$ represent the latent embeddings from different kinds of EEG signals (e.g., TMR-related and image-evoked EEG signals). Besides, $\bm{h}$ can also be the representations from other modalities (e.g., audio representations from wav2vec 2.0 \cite{defossez2023decoding} or image representations from CLIP \cite{song2023decoding}) in the ablation study.

\paragraph{Shared Encoder.}In our experiments, directly aligning the neural representation sequence of TMR-related EEG signals ($\mathcal{X}^{tmr}$) with that of both $\mathcal{X}^{img}$ and $\mathcal{X}^{aud}$ often leads to a slight degradation, compared to aligning with either of them individually. The initial design demands the model to simultaneously find the optimal alignment among them, which poses a challenging task. To further smooth the aligning process, we feed the output embeddings $\mathcal{E}^{tmr}$ from $\mathcal{X}^{tmr}$ into the transformer encoders of awake signals, yielding $\mathcal{E}^{tmr2img}$ and $\mathcal{E}^{tmr2aud}$ respectively. Therefore, we utilize the transformed embeddings to calculate contrastive loss, i.e., $\mathcal{E}^{img}$ vs. $\mathcal{E}^{tmr2img}$ and $\mathcal{E}^{aud}$ vs. $\mathcal{E}^{tmr2aud}$.

Besides, all of these embeddings $\langle \mathcal{E}^{tmr},\mathcal{E}^{tmr2img},\mathcal{E}^{tmr2aud}\rangle$ are utilized to predict the semantic labels, and averaged to get the final prediction $\bm{y}$, which resembles the model ensemble method. This method enables fine-tuning the pre-trained model on the awake dataset of that subject, which improves the decoding performance of REM sleep (as illustrated in Table \ref{table:result-awake-finetune}). The model (w/o awake data \& contrastive loss) serves as the baseline SI-SD; see Appendix \ref{sec:supp-model-details} for detailed definitions.

\paragraph{Training Objective.}The total loss for training SI-SD consists of two parts: (1) classification loss, and (2) contrastive loss. For each data item $\langle \mathcal{X}^{img},\mathcal{X}^{aud},\mathcal{X}^{tmr},\bm{y}\rangle$, all of these transformed embeddings $\langle \mathcal{E}^{tmr},\mathcal{E}^{img},\mathcal{E}^{aud},\mathcal{E}^{tmr2img},\mathcal{E}^{tmr2aud}\rangle$ are utilized to calculate the classification loss:
\begin{equation}
  \mathcal{L}_{cls}=\mathcal{L}_{cls}^{tmr}+\mathcal{L}_{cls}^{img}+\mathcal{L}_{cls}^{aud} +\mathcal{L}_{cls}^{tmr2img}+\mathcal{L}_{cls}^{tmr2aud}.
\end{equation}
And the contrastive loss is calculated as demonstrated before:
\begin{equation}
  \mathcal{L}_{contra}=\mathcal{L}_{contra}^{tmr2img}+\mathcal{L}_{contra}^{tmr2aud} +\mathcal{L}_{contra}^{img2aud}.
\end{equation}
Therefore, the total training objective of SI-SD is:
\begin{equation}
  \mathcal{L}_{total}=\mathcal{L}_{cls}+\mathcal{L}_{contra}.
\end{equation}

\section{Experiments}
\subsection{Dataset}
Due to the lack of datasets related to sleep semantic decoding tasks, we designed an innovative cognitive neuroscience experiment and collected a well-annotated sleep semantic decoding EEG dataset (including 134 healthy subjects). The dataset is recorded at a rate of 500 Hz using a 64-channel EEG cap. Notably, the dataset from 122 subjects is collected in one laboratory, whereas the dataset from the remaining 12 subjects is collected in a different laboratory. The total dataset contains $\sim$1000 hours of sleep recordings and $\sim$500 hours of awake recordings. See Appendix \ref{sec:supp-expr-design} for more details.

\paragraph{Image-Evoked Dataset.}For each subject, $\sim$1.8 hours of EEG recordings are image-evoked EEG signals. The EEG signals are segmented into about 2000 (averaged across subjects) 0.8-second samples from 0s to 0.8s according to the \textbf{onset} of the image stimulus, each of which is paired with the corresponding semantic label (from 15 pre-determined semantic concepts).

\paragraph{Audio-Evoked Dataset.}For each subject, $\sim$1.8 hours of EEG recordings are audio-evoked EEG signals. The EEG signals are segmented into about 2000 0.8-second samples from 0s to 0.8s according to the \textbf{onset} of the audio stimulus, each paired with the corresponding semantic label.

\paragraph{TMR-Related Dataset.}For each subject, $\sim$7.2 hours of EEG recordings are TMR-related EEG signals. After sleep staging with YASA \cite{vallat2021open}, the EEG signals are segmented into $\sim$1200 2-second NREM 2/3 samples and $\sim$300 2-second REM samples from -0.2s to 1.8s according to the \textbf{offset} of the audio stimulus, each paired with the corresponding semantic label. As NREM 2/3 and REM are distinct sleep stages characterized by different neural patterns, we split these samples to train NREM 2/3 and REM sleep semantic decoders separately.

\subsection{Implementation Details}
\label{sec:imple-details}
\paragraph{Preprocess.}We first filter the EEG signals between 0.1Hz and 40Hz to remove low-frequency noise. Then, a notch filter of 50Hz is applied to avoid power-line interference. After that, all EEG signals are resampled to 100Hz and average re-referenced \cite{yao2019reference}. Finally, we perform z-score normalization on each channel to guarantee normalized data scales across all channels.

\paragraph{Model Configurations.}The "Spatial Encoder" contains three residual convolution blocks and one linear projection, transforming the original EEG signals into token embeddings with $d=128$. The following "Transformer Encoder" contains a multi-layer Transformer encoder (12-layer for TMR-related EEG signals; shorter for other kinds of EEG signals) with model dimension $d=128$, inner dimension (FFN) $d_{ff}=1024$ and 6 attention heads. See Appendix \ref{sec:supp-model-details} for more details.

\paragraph{Model Training.}As mentioned before, the whole dataset comes from two different laboratories. According to their sources, we split the set of subjects $\mathcal{S}$ (including 134 subjects) into two subsets: $\mathcal{S}_{1}$ (including 122 subjects) and $\mathcal{S}_{2}$ (including 12 subjects). We utilize the datasets $\{\langle \mathcal{D}^{img}_{s},\mathcal{D}^{aud}_{s},\mathcal{D}^{tmr}_{s}\rangle\}_{s\in\mathcal{S}_{1}}$ from $\mathcal{S}_{1}$ to train models. For the training of the NREM 2/3 sleep decoder, there are approximately 240,000 image-evoked data, 240,000 audio-evoked data, and 120,000 TMR-related NREM 2/3 sleep data across 122 subjects. For the training of the REM sleep decoder, there are approximately 30,000 TMR-related REM sleep data along with the aforementioned awake data. The model undergoes supervised training on a Linux system with 2 CPUs (Intel Xeon Gold 6230 40-Core Processor) and 8 GPUs (NVIDIA Tesla V100 32GB) for $\sim$2 days. Then, we evaluate the performance of the model on the remaining datasets $\{\langle \mathcal{D}^{img}_{s},\mathcal{D}^{aud}_{s},\mathcal{D}^{tmr}_{s}\rangle\}_{s\in\mathcal{S}_{2}}$ from $\mathcal{S}_{2}$.

\paragraph{Subject-wise Fine-tuning.}During the downstream fine-tuning, we split the dataset into training, validation, and testing splits with a size roughly proportional to 80\%, 10\%, and 10\%. We utilize data augmentation, as described in Appendix \ref{sec:supp-data-augmentation}, to make the most of the gathered dataset. To ensure a fair comparison, all models are fine-tuned on the TMR-related sub-dataset $\mathcal{D}^{tmr}_{s}$. Our experiments are conducted on one V100 GPU by Python 3.8.15 and TensorFlow 2.6.0 + CUDA 11.2. The best models are trained based on the training set, selected from the validation set, and finally evaluated on the test set. For model comparison, we report the average and standard error values (of all subjects) on six different random seeds to obtain comparable results. For the results of the subject-wise evaluation, we report the average and standard deviation values (of each subject) in Appendix \ref{sec:supp-subj-evaluation}.

\subsection{Comparison with Other Models}
Table \ref{table:result-main-avg} presents the results of our SI-SD model and other supervised baselines, with the best in \textbf{bold} and the second \underline{underlined}. Our SI-SD model consistently outperforms all baselines. Besides, we further align the neural representation sequence of TMR-related EEG signals with that of other modalities; see Appendix \ref{sec:supp-model-details} for detailed definitions. Due to the limited semantic categories, treating the entire neural representation sequence as the basic units leads to the degradation of contrastive loss into classification loss. Therefore, "SI-SD+CLIP" performs similarly to the baseline SI-SD model.

In comparison, decomposing the entire neural representation sequence into neural representation units can provide more fine-grained alignment. As we expected, "SI-SD+wav2vec" outperforms both "SI-SD (baseline)" and "SI-SD+CLIP". However, directly aligning neural representation sequences across trials for the same category \cite{cho2023neural} within the sleep dataset, i.e., "SI-SD+NLA", performs slightly worse than "SI-SD+wav2vec", especially on NREM 2/3 sleep dataset. We hypothesize that this result is because external stimuli evoke the semantic components with varying degrees of intensity during NREM 2/3 sleep, which makes the alignment harder; see Section \ref{sec:so-analysis} for further analysis.

As demonstrated in Appendix \ref{sec:supp-expr-design}, the audio stimulus is natural sounds, e.g., the cry of sheep, instead of human speech. However, wav2vec 2.0 \cite{baevski2020wav2vec} is pre-trained on a large corpus of human speech, whose representations might not adequately capture the semantic processing of the audio stimulus. In comparison, the neural representation sequence from the awake dataset can not only provide a template for alignment but also capture the semantic processing of the same brain. "SI-SD+Awake-Data Guide" attains zero-shot\footnote{We refer to the decoding performance on unseen subjects as zero-shot decoding performance, as all subjects share 15 pre-determined semantic concepts.} decoding performances of 24.12\% and 20.27\% on NREM 2/3 and REM sleep decoding tasks, respectively, surpassing "SI-SD+wav2vec" by 1.37\% and 2.38\% respectively.

\begin{table}[h]
  \caption{The performance of different methods in different experiment settings. For each method, the 15-way classification accuracy (\%) along with the standard error (\%) are reported.}
  \label{table:result-main-avg}
  \centering
  \begin{tabular}{lccccc}
    \toprule
    \multirow{2}{*}{\textbf{Methods}}  & \multicolumn{2}{c}{\textbf{Zero-Shot}} & \multicolumn{2}{c}{\textbf{Fine-tune}} \\
    \cmidrule(lr){2-3}\cmidrule(lr){4-5}
    & NREM 2/3 & REM & NREM 2/3 & REM \\
    \midrule
    Lasso-GLM\cite{horikawa2013neural} & 11.20$\pm$0.37 & 11.38$\pm$0.71 & 13.74$\pm$0.35 & 17.11$\pm$1.12 \\
    \midrule
    EEG-Net\cite{lawhern2018eegnet}    & 17.30$\pm$0.37 & 14.76$\pm$1.11 & 19.78$\pm$0.66 & 20.89$\pm$1.95 \\
    EEG-Conformer\cite{song2022eeg}    & 19.73$\pm$0.48 & 14.96$\pm$0.79 & 24.24$\pm$0.83 & \underline{24.05$\pm$1.82} \\
    \midrule
    SI-SD (single-subject)           & - & - & 20.11$\pm$0.86 & 12.31$\pm$0.98 \\
    \midrule
    SI-SD (baseline)                 & 21.01$\pm$0.41 & 16.72$\pm$1.17 & 26.47$\pm$0.55 & 26.46$\pm$2.02 \\
    +CLIP\cite{song2023decoding}       & 21.02$\pm$0.38 & 16.18$\pm$1.04 & 26.22$\pm$0.47 & 26.48$\pm$1.99 \\
    +NLA\cite{cho2023neural}           & 21.54$\pm$0.47 & 17.43$\pm$1.07 & 27.09$\pm$0.50 & 27.72$\pm$2.03 \\
    +wav2vec\cite{defossez2023decoding}& \underline{22.75$\pm$0.55} & \underline{17.89$\pm$1.10} & \underline{28.87$\pm$0.53} & \underline{29.19$\pm$2.16} \\
    +Awake-Data Guide                  & \textbf{24.12$\pm$0.67} & \textbf{20.27$\pm$1.41} & \textbf{30.32$\pm$0.70} & \textbf{30.02$\pm$2.22} \\
    \bottomrule
  \end{tabular}
\end{table}

\subsection{Results with Awake-Data Fine-tuning}
Since our model shares the transformer encoders, in the pathways for awake signals, with sleep signals, we can utilize the awake dataset of that subject to further enhance sleep semantic decoding in both zero-shot and fine-tuning subject-wise evaluations. Specifically, in the zero-shot subject-wise evaluation, we utilize the awake dataset to fine-tune the transformer encoders in the pathways for awake signals; in the fine-tuning subject-wise evaluation, we fine-tune the whole model with the awake and sleep datasets. Table \ref{table:result-awake-finetune} demonstrates the fine-tuning results based on the "SI-SD+Awake-Data Guide" model. As the neural patterns during wakefulness and REM sleep are relatively similar, awake-data fine-tuning mainly improves the sleep semantic decoding of REM sleep.

\begin{table}[h]
  \caption{The performance with or without awake-data fine-tuning.}
  \label{table:result-awake-finetune}
  \centering
  \begin{tabular}{lcccc}
    \toprule
    \multirow{2}{*}{\textbf{Methods}}  & \multicolumn{2}{c}{\textbf{Zero-Shot}} & \multicolumn{2}{c}{\textbf{Fine-tune}} \\
    \cmidrule(lr){2-3}\cmidrule(lr){4-5}
    & NREM 2/3 & REM & NREM 2/3 & REM \\
    \midrule
    SI-SD (w/o awake-data fine-tuning) & 24.12$\pm$0.67 & 20.27$\pm$1.41 & 30.32$\pm$0.70 & 30.02$\pm$2.22 \\
    SI-SD (w/ awake-data fine-tuning)  & 24.20$\pm$0.56 & \textbf{21.39$\pm$1.40} & 30.51$\pm$0.62 & \textbf{31.65$\pm$1.88} \\
    \bottomrule
  \end{tabular}
\end{table}

\subsection{Analysis related to Slow Oscillation Event}
\label{sec:so-analysis}
Previous neuroscience studies \cite{schreiner2021endogenous,ngo2022shaping,schreiner2023respiration} demonstrate that triple-coupling (i.e., the coupling of Slow Oscillation (SO), spindle, ripple) is crucial for memory consolidation and "SO" event greatly impacts decoding performance during NREM 2/3 sleep. However, due to EEG recording constraints, we only examine the effects of SO-spindle coupling on sleep semantic decoding using our collected dataset.

\begin{figure}[h]
  \centering
  \includegraphics[width=\linewidth]{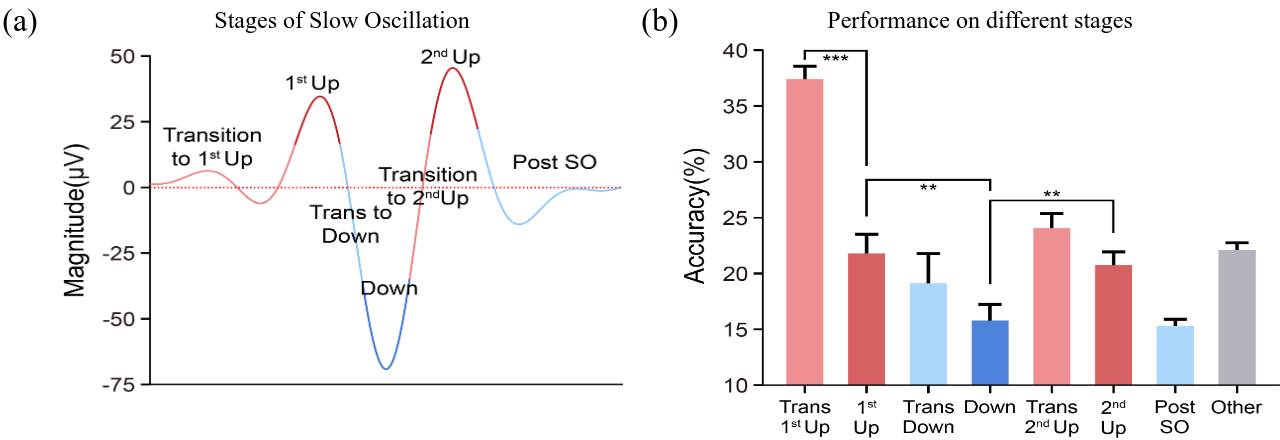}
  \caption{\textbf{The analysis related to the "Slow Oscillation" event.} \textbf{(a).} The stages of the "Slow Oscillation" event. We divide NREM 2/3 sleep samples into 8 groups according to the onset of their audio stimuli. \textbf{(b).} The decoding performance of different SO-groups during NREM 2/3 sleep.}
  \label{fig:slow-oscillation}
\end{figure}

In our collected dataset, the "SO" events are detected in about 40\% of the samples during NREM 2/3 sleep. As illustrated in Figure \ref{fig:slow-oscillation}(a), we divide these samples into 8 non-overlapping groups according to the onset of audio stimuli; see Appendix \ref{sec:supp-so-analysis} for detailed operational definitions. Then, we evaluate the zero-shot decoding performance of the "SI-SD+wav2vec" model in these SO-groups. The decoding performance of the "1st-Up" and "2nd-Up" groups is significantly higher than that of the "Down" group, which validates the previous neuroscience findings \cite{schreiner2021endogenous}. Besides, we uncover new features -- the decoding performance on the "Trans-1st-Up" group ($\sim$10\% of the entire NREM 2/3 sleep dataset) achieves 37.41\%, as illustrated in Table \ref{table:result-n2n3-trans1up}. By incorporating the awake dataset during supervised training, "SI-SD+Awake-Data Guide" achieves 40.02\% zero-shot performance, showcasing a greater improvement compared to the entire NREM 2/3 sleep dataset. This result further demonstrates the effectiveness of aligning neural latent sequence position-wise between wakefulness and sleep. Due to the limited sample size, the fine-tuning results are not presented.

\begin{table}[h]
  \caption{The zero-shot performance of different methods on the "Trans-1st-Up" group.}
  \label{table:result-n2n3-trans1up}
  \centering
  \begin{tabular}{ccccc}
    \toprule
    \textbf{Methods} & Lasso-GLM & SI-SD (baseline) & +wav2vec\cite{defossez2023decoding} & +Awake-Data Guide \\
    \midrule
    - & 23.00$\pm$0.79 & 33.58$\pm$1.19 & \underline{37.41$\pm$1.10} & \textbf{40.02$\pm$1.04} \\
    \bottomrule
  \end{tabular}
\end{table}

Besides, we further examine the effect of spindle events within them separately. Our findings reveal that the "w/ spindle" sub-group (i.e., with SO-spindle coupling) outperforms the "w/o spindle" sub-group, as shown in Figure \ref{fig:spindle-performance}. Notably, the "Trans-1st-Up" group has the highest proportion of SO-spindle coupling ($>50\%$), partially explaining its highest decoding performance. Since spindle event detection relies on the amplitude of specific frequency bands, we observe that decoding performance correlates positively with the amplitude of these bands in the "Trans-1st-Up" group. According to the temporal-frequency analysis, the amplitude of other frequency bands is also improved in these SO-spindle coupling samples, as shown in Figure \ref{fig:so-spindle-statistics}(b). Hence, we attribute the improved amplitude to increased semantic components and hypothesize that it relates to enhanced memory content.

To ensure that awake and sleep signals indeed share semantic information, we train Lasso GLMs \cite{horikawa2013neural} on awake datasets and evaluate them on these SO-groups. Table \ref{table:result-transfer} shows that the Lasso GLM trained on the awake-image dataset achieves 17.98\% on the "Trans-1st-Up" group, indicating shared semantic components in EEG signals during wakefulness and NREM 2/3 sleep. Due to the limited sample size, the transfer results from NREM 2/3 sleep groups to the awake datasets are not presented.

\begin{table}[h]
  \caption{The transfer performance from different sources on the NREM 2/3 sleep dataset.}
  \label{table:result-transfer}
  \centering
  \begin{threeparttable}
  \begin{tabular}{lcccccccc}
    \toprule
    \textbf{Experiments\footnotemark[1]} & Trans-1st-Up & 1st-Up & Trans-Down & Down \\
    \midrule
    $\mathcal{X}^{img}\rightarrow\mathcal{X}^{tmr}$ & \textbf{17.98$\pm$0.86} & 13.37$\pm$0.42 & 12.83$\pm$0.32 & 11.99$\pm$0.28 \\
    $\mathcal{X}^{aud}\rightarrow\mathcal{X}^{tmr}$ & \textbf{18.67$\pm$0.97} & 13.23$\pm$0.54 & 12.60$\pm$0.38 & 12.01$\pm$0.33 \\
    \midrule
    - & Trans-2nd-Up & 2nd-Up & Post-SO & Other \\
    \midrule
    $\mathcal{X}^{img}\rightarrow\mathcal{X}^{tmr}$ & 13.17$\pm$0.65 & 12.85$\pm$0.40 & 12.06$\pm$0.41 & 10.36$\pm$0.25 \\
    $\mathcal{X}^{aud}\rightarrow\mathcal{X}^{tmr}$ & 13.24$\pm$0.66 & 12.64$\pm$0.61 & 11.89$\pm$0.38 & 10.37$\pm$0.18 \\
    \bottomrule
  \end{tabular}
  \begin{tablenotes}
    \item[1] The chance-level decoding performance is $\sim$6.67\%.
  \end{tablenotes}
  \end{threeparttable}
\end{table}

\section{Limitations}
\label{sec:limitations}
Despite SI-SD’s enhancements in sleep semantic decoding with awake-guided neural latent alignment, it is still restricted to the close-set sleep semantic decoding tasks (i.e., the concept set only includes 15 pre-determined concepts). Given the constraints of human cognitive load, it's challenging to ensure the quality of the dataset (w/ numerous semantic categories) collected within one night. Besides, since current neuroscience studies \cite{turker2022behavioral,liu2022decoding} also utilize the limited concept set to study sleep, 15 concepts during sleep are already enough for them. Furthermore, as our work marks the initial endeavor in the field of sleep semantic decoding, this dataset serves as an effective benchmark for developing models capable of accurately extracting semantic information from noisy sleep EEG signals.

Additionally, our proposed method is limited to supervised methods. Recent EEG decoding studies \cite{yang2024biot,jiang2024large} learn general embeddings through self-supervision. LaBraM \cite{jiang2024large} builds univariate representations and learns to decouple the confused information within each channel through vector-quantized spectrum prediction. Besides, the key challenge observed in sleep semantic decoding is that external stimuli evoke semantic components with varying degrees of intensity, contributing to the presence of noisy EEG signals. We hope that with discrete codebook-guided pre-training, we can better decouple semantic components from noisy EEG signals, enhancing sleep semantic decoding.

\section{Conclusion}
This paper proposes the SI-SD model for sleep semantic decoding, aligning neural latent sequence position-wise between wakefulness and sleep to enhance performance. To evaluate our model, we collect a well-annotated sleep semantic decoding EEG dataset to address the lack of datasets related to sleep semantic decoding. Comprehensive experiments demonstrate that our model outperforms all baselines, and awake signals provide a better template to extract semantic components from sleep signals. Moreover, we analyze the effect of "Slow Oscillation" events during NREM 2/3 sleep on decoding performance, validating existing neuroscience findings and uncovering new features. Together, we hope our work will influence future EEG-based sleep decoding models with more consideration over how to extract semantic components from noisy sleep EEG signals effectively.

\clearpage

\subsubsection*{Ethics Statement}
\label{sec:ethics-state}
Experiments that contribute to this work were approved by IRB. All subjects consent to participate and are compensated with cash.

\subsubsection*{Reproducibility Statement}
\label{sec:reprod-state}
Code to train models and reproduce the results is submitted as part of the supplementary materials and can be accessed here: \href{TODO}{TODO}, including a demo dataset of 3 subjects for downstream fine-tuning. The whole dataset will be released upon publication.

\bibliography{neurips_2024}
\bibliographystyle{plain}

\clearpage
\appendix
\section{Experiment Design}
\label{sec:supp-expr-design}
As demonstrated in Fig.\ref{fig:expr-design}, our experiment paradigm contained three stages:
\begin{itemize}
  \item pre-sleep function localizer stage,
  \item overnight sleep target memory reactivation stage,
  \item post-sleep function localizer stage.
\end{itemize}
First, we selected 15 semantic classes: alarm, apple, ball, book, box, chair, kiwi, microphone, motorcycle, pepper, sheep, shoes, strawberry, tomato, watch. Then, we used the corresponding pictures and sounds to form the set of image stimuli (including 15 different pictures) and the set of audio stimuli (including 15 different sounds). It's worth noting that the sound corresponding to the picture can easily trigger the recall of the associated picture, e.g., the picture of "sheep" is paired with the sound of the "sheep" instead of other unrelated sounds.

\subsection{Pre\&Post-Sleep Function Localizer Stage}
In the pre-sleep function localizer stage, the subject was instructed to perform two tasks: the image-audio task and the audio-image task. In the image-audio task, the subject was presented with 600 trials, with each trial containing one image stimulus and one audio stimulus, in image-audio order. We randomly selected one image stimulus and one audio stimulus from the set of image stimuli and the set of audio stimuli, respectively, where the image stimulus and the audio stimulus may be inconsistent (i.e., coming from different semantic classes). In total, we had 210 inconsistent trials and 390 consistent trials. The purpose of this design is solely to confirm that participants have formed correct and stable stimulus pair memories.

\begin{figure}[h]
  \centering
  \includegraphics[width=\linewidth]{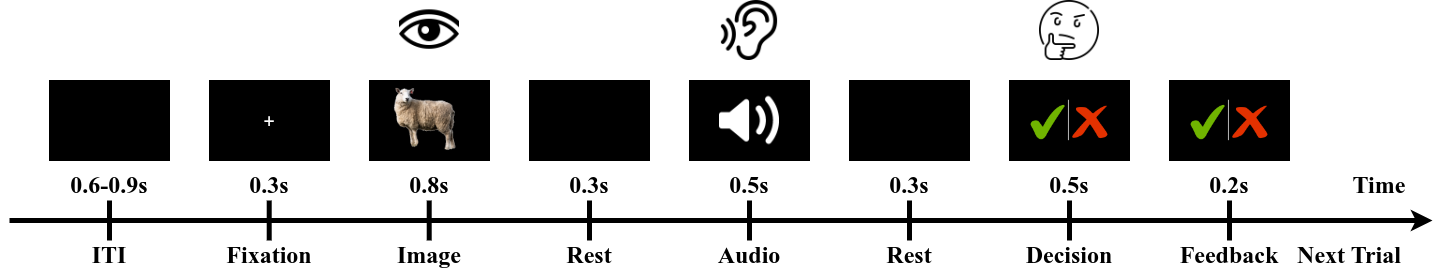}
  \caption{The experiment pipeline of each trial during the function localizer stage. Here, we only illustrate the pipeline for the image-audio task. For the audio-image task, we simply reverse the presentation order of the image and audio stimuli.}
  \label{fig:expr-trial}
\end{figure}

For each trial, the subject was presented with the following items, as illustrated in Figure \ref{fig:expr-trial}:
\begin{itemize}
  \item a blank screen for 0.6s-0.9s (i.e., inter-trial interval),
  \item a cross centered at the screen for 0.3s,
  \item an image stimulus for 0.8s,
  \item a blank screen for 0.3s,
  \item an audio stimulus for 0.5s,
  \item a blank screen for 0.3s.
\end{itemize}
Then, the subject was required to decide whether the presented image stimulus and audio stimulus were matched (i.e., coming from the same semantic class). Finally, we presented the behavior feedback for 0.2s. In total, the image-audio task took about 48 minutes.

The setup of the audio-image task was similar to that of the image-audio task. The only difference between them was that we presented image-audio pairs in audio-image order.

In the post-sleep function localizer stage, the subject was required to execute the same tasks in the pre-sleep function localizer stage.

\subsection{Overnight Sleep Target Memory Reactivation Stage}
In the overnight sleep target memory reactivation stage, we utilized a closed-loop stimulation system that allows for real-time, automatic sleep staging. As the subject reached the N2/3 stage of NREM sleep, the subject was presented with audio stimuli (which are the same as those during wakefulness, each lasting 0.5s), randomly selected from the set of audio stimuli, every 4-6 seconds. This approach provided clear timing and content of memory reactivation during sleep, aiding in the training of a neural decoder on these cued sleep intervals.

\subsection{The Statistics of the Dataset}
\begin{figure}[h]
  \centering
  \includegraphics[width=\linewidth]{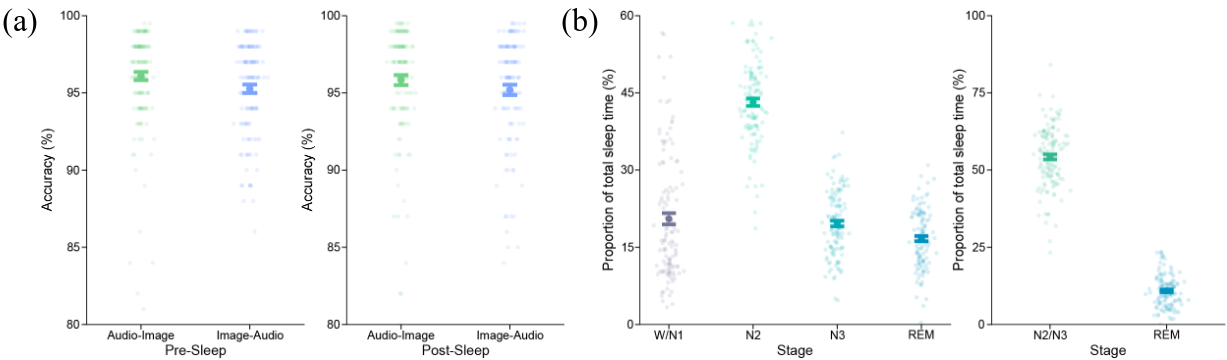}
  \caption{The statistics of the collected dataset.}
  \label{fig:dataset-statistics}
\end{figure}

To guarantee the quality of the awake dataset, we tallied the rates of behavioral correctness for each subject, as depicted in Figure \ref{fig:dataset-statistics}(a). Figure \ref{fig:dataset-statistics}(b) illustrates the distribution of each sleep stage within the sleep dataset.

In summary, the collected dataset includes 134 healthy subjects (122 subjects from one laboratory and 12 subjects from another). The total dataset contains:
\begin{itemize}
  \item $\sim$1000 hours sleep EEG recordings.
  \item $\sim$500 hours awake EEG recordings.
\end{itemize}
For different tasks, the total dataset contains:
\begin{itemize}
  \item $\sim$270,000 0.8-second image-evoked EEG samples.
  \item $\sim$270,000 0.8-second audio-evoked EEG samples.
  \item $\sim$135,000 2-second TMR-related NREM 2/3 sleep EEG samples.
  \item $\sim$35,000 2-second TMR-related REM sleep EEG samples.
\end{itemize}
For each subject, the dataset contains (on average):
\begin{itemize}
  \item $\sim$2000 0.8-second image-evoked EEG samples.
  \item $\sim$2000 0.8-second audio-evoked EEG samples.
  \item $\sim$1200 2-second TMR-related NREM 2/3 sleep EEG samples.
  \item $\sim$300 2-second TMR-related REM sleep EEG samples.
\end{itemize}

\clearpage
\section{Details of Baselines}
\label{sec:supp-baseline-details}
In experiments, we compare our model to the existing supervised methods on brain signals. The details of these baseline models are given here:
\begin{itemize}
  \item \textbf{Lasso-GLM}\cite{horikawa2013neural}: A linear classifier with lasso regularization, which is commonly used in neuroscience research. As previous studies \cite{horikawa2013neural,turker2022behavioral,schreiner2021endogenous} mainly utilize this model to decode semantic contents during sleep, this model is suitable to serve as a baseline for comparison.
  \item \textbf{EEG-Net}\cite{lawhern2018eegnet}: A supervised model that utilizes the depthwise and separable convolutions to construct an EEG-specific model, which encapsulates well-known EEG feature extraction concepts for BCI. Since this model is robust enough to learn a wide variety of interpretable features over a range of BCI tasks, it is suitable as a baseline for comparison.
  \item \textbf{EEG-Conformer}\cite{song2022eeg}: A supervised model that consists of both CNN module and Transformer module to encapsulate local and global features in a unified EEG classification framework. Since this model is widely used in the field of EEG decoding, it is suitable as a baseline for comparison.
\end{itemize}
When evaluating the decoding performance of these baseline models, we follow the same experiment setup of the SI-SD model:
\begin{itemize}
  \item We utilize the sleep dataset from 122 subjects in $\mathcal{S}_{1}$ to train the model and conduct subject-wise zero-shot evaluation.
  \item For subject-wise fine-tuning, we split the sleep dataset of that subject into training, validation, and testing splits with a size roughly proportional to 80\%, 10\%, and 10\%.
  \item The data samples are resampled to the specified sampling rate of each model.
\end{itemize}

\section{Model Details}
\label{sec:supp-model-details}
\subsection{The Architecture of Neural Transformer}
The architecture of the "Neural Transformer", as illustrated in Figure \ref{fig:neural-xfmr}, contains two parts: (1) Spatial Encoder, and (2) Transformer Encoder. The hyperparameters of "Neural Transformer" are shown in Table \ref{table:model-neural-xfmr}. Since there are different numbers of Transformer layers on different pathways of the SI-SD model, we don't specify the number of Transformer layers in the "Transformer Encoder" here.
\begin{table}[h]
  \caption{The hyperparameters of "Neural Transformer" model.}
  \label{table:model-neural-xfmr}
  \centering
  \begin{threeparttable}
  \begin{tabular}{cccc}
    \toprule
    \textbf{Module} & \textbf{Name} & \textbf{Value} \\
    \midrule
    \multirow{7}{*}{Spatial Encoder} & \# of Input Channels & \{55,256,256,256,256,256\} \\
    & \# of Output Channels & \{256,256,256,256,256,256\} \\
    & Kernel Size & \{9,9,9,9,9,9\} \\
    & Dilation Rate & \{1,2,4,1,2,4\} \\
    & Stride & \{1,1,1,1,1,1\} \\
    & Padding & "same" \\
    & Linear Projection & $256\rightarrow 256\rightarrow 128$ \\
    \midrule
    \multirow{7}{*}{Transformer Encoder} & \# of Transformer Layers & - \\
    & Hidden Size & 128 \\
    & MLP Size & 1024 \\
    & MLP Dropout Ratio & \{0.5,0.5\} \\
    & \# of Attention Heads & 6 \\
    & Attention Head Size & 64 \\
    & Linear Projection\footnotemark[1] & $128\rightarrow 256$ \\
    \bottomrule
  \end{tabular}
  \begin{tablenotes}
    \item[1] Linear projection is a position-wise projection after the "Transformer Encoder".
  \end{tablenotes}
  \end{threeparttable}
\end{table}

\subsection{The Architecture of SI-SD model}
The overall architecture of our SI-SD model is illustrated in Figure \ref{fig:sisd}, which contains three pathways for different EEG signals: (1) TMR-related EEG signals $\mathcal{X}^{tmr}$ in the middle section, (2) audio-evoked EEG signals $\mathcal{X}^{aud}$ in the upper section, and (3) image-evoked EEG signals $\mathcal{X}^{img}$ in the lower section. The hyperparameters for SI-SD training are shown in Table \ref{table:model-sisd}. Notably, the "Classification Head" is individually instantiated for three pathways:
\begin{itemize}
  \item $\mathcal{E}^{tmr}$ is fed into the "Classification Head" in the pathway for $\mathcal{X}^{tmr}$.
  \item $\langle \mathcal{E}^{aud},\mathcal{E}^{tmr2aud}\rangle$ are separately fed into the "Classification Head" in the pathway for $\mathcal{X}^{aud}$.
  \item $\langle \mathcal{E}^{img},\mathcal{E}^{tmr2img}\rangle$ are separately fed into the "Classification Head" in the pathway for $\mathcal{X}^{img}$.
\end{itemize}
In implementation, the predictions from $\langle \mathcal{E}^{tmr},\mathcal{E}^{tmr2aud},\mathcal{E}^{tmr2img}\rangle$ are averaged to get the final prediction of sleep semantic decoding.
\begin{table}[h]
  \caption{The hyperparameters for SI-SD training.}
  \label{table:model-sisd}
  \centering
  \begin{threeparttable}
  \begin{tabular}{cccc}
    \toprule
    \textbf{Module} & \textbf{Sub-Module} & \textbf{Name} & \textbf{Value} \\
    \midrule
    \multirow{1}{*}{Pathway for $\mathcal{X}^{tmr}$} & Neural Transformer & \# of Transformer Layers & 12 \\
    \midrule
    \multirow{1}{*}{Pathway for $\mathcal{X}^{aud}$} & Neural Transformer & \# of Transformer Layers & 6 \\
    \midrule
    \multirow{1}{*}{Pathway for $\mathcal{X}^{img}$} & Neural Transformer & \# of Transformer Layers & 4 \\
    \midrule
    \multirow{9}{*}{Contrastive Block} & \multirow{3}{*}{$\mathcal{E}^{tmr}$ vs. $\mathcal{E}^{aud}$} & Linear Projection of $\mathcal{E}^{tmr}$ & $256\rightarrow 32$ \\
    & & Linear Projection of $\mathcal{E}^{aud}$ & $256\rightarrow 80$ \\
    & & Flatten & - \\
    \cline{2-4}
    & \multirow{3}{*}{$\mathcal{E}^{tmr}$ vs. $\mathcal{E}^{img}$} & Linear Projection of $\mathcal{E}^{tmr}$ & $256\rightarrow 32$ \\
    & & Linear Projection of $\mathcal{E}^{img}$ & $256\rightarrow 80$ \\
    & & Flatten & - \\
    \cline{2-4}
    & \multirow{3}{*}{$\mathcal{E}^{aud}$ vs. $\mathcal{E}^{img}$} & Linear Projection of $\mathcal{E}^{aud}$ & $256\rightarrow 64$ \\
    & & Linear Projection of $\mathcal{E}^{img}$ & $256\rightarrow 64$ \\
    & & Flatten & - \\
    \midrule
    \multirow{2}{*}{Classification Head\footnotemark[1]} & \multirow{2}{*}{-} & Flatten & - \\
    & & Linear Projection & $\mathrm{seq\_len}\times 256\rightarrow 128 \mathrm{(RELU)} \rightarrow 15$ \\
    \midrule
    \multirow{6}{*}{\makecell[c]{Optimizer}} & \multirow{6}{*}{-} & Batch Size & 512 (64 for fine-tuning) \\
    & & Learning Rate & 8e-5 (2e-5 for fine-tuning) \\
    & & Optimizer Type & AdamW \\
    & & Adam $\beta$ & $(0.9,0.99)$ \\
    & & Weight Decay & 2e-5 \\
    & & Total Epochs & 100 \\
    \bottomrule
  \end{tabular}
  \begin{tablenotes}
    \item[1] Classification Head is individually instantiated for three pathways.
  \end{tablenotes}
  \end{threeparttable}
\end{table}

In the ablation experiments shown in Table \ref{table:result-main-avg}, our models have different suffixes:
\begin{itemize}
  \item \textbf{SI-SD (baseline):} This model consists of the entire pathway of the sleep data and two transformer encoders in the pathway of the awake data. The embeddings $\langle \mathcal{E}^{tmr},\mathcal{E}^{tmr2img},\mathcal{E}^{tmr2aud}\rangle$ from all pathways are utilized to predict the semantic labels, and averaged to get the final prediction $\bm{y}$. This model undergoes supervised training using the sleep dataset from 122 subjects in $\mathcal{S}_{1}$, followed by subject-wise zero-shot evaluation and fine-tuning using the sleep dataset from 12 subjects in $\mathcal{S}_{2}$.
  \item \textbf{SI-SD (single-subject):} The architecture of this model is the same as the "SI-SD (baseline)" model. Without supervised training on $\mathcal{S}_{1}$, this model is directly trained on the sleep dataset of each subject in $\mathcal{S}_{2}$.
  \item \textbf{SI-SD + CLIP:} The architecture of this model is the same as the "SI-SD (baseline)" model. Besides, during the supervised training process over $\mathcal{S}_{1}$, we project the embedding sequences $\langle \mathcal{E}^{tmr},\mathcal{E}^{tmr2img},\mathcal{E}^{tmr2aud}\rangle$ from all pathways into representation units separately. Then, we align them with the representation of the corresponding image from CLIP \cite{radford2021learning}, following the operations outlined in NICE \cite{song2023decoding}.
  \item \textbf{SI-SD + NLA:} The architecture of this model is the same as the "SI-SD (baseline)" model. Besides, during the supervised training process over $\mathcal{S}_{1}$, we align the embedding sequences $\langle \mathcal{E}^{tmr},\mathcal{E}^{tmr2img},\mathcal{E}^{tmr2aud}\rangle$ across trials for the same category, following the operations outlined in NLA \cite{cho2023neural}.
  \item \textbf{SI-SD + wav2vec:} The architecture of this model is the same as the "SI-SD (baseline)" model. Besides, during the supervised training process over $\mathcal{S}_{1}$, we align the embedding sequences $\langle \mathcal{E}^{tmr},\mathcal{E}^{tmr2img},\mathcal{E}^{tmr2aud}\rangle$ from all pathways with the representation sequence of the corresponding audio from wav2vec 2.0 \cite{baevski2020wav2vec}, following the operations proposed by D$\mathrm{\acute{e}}$fossez et al. \cite{defossez2023decoding}.
  \item \textbf{SI-SD + Awake-Data Guide:} The complete model illustrated in Figure \ref{fig:sisd}. This model undergoes supervised training using both the awake and sleep datasets from 122 subjects in $\mathcal{S}_{1}$, followed by subject-wise zero-shot evaluation and fine-tuning using the sleep dataset from 12 subjects in $\mathcal{S}_{2}$. The performance with or without awake-data fine-tuning is demonstrated in Table \ref{table:result-awake-finetune}.
\end{itemize}

\clearpage
\section{Data Augmentation}
\label{sec:supp-data-augmentation}
We apply data augmentation in the sleep dataset to enhance the robustness of learned representations for sleep semantic decoding. Since the awake samples only last for 0.8 seconds, we don't apply data augmentation to the awake dataset.

As demonstrated in Appendix \ref{sec:supp-expr-design}, the intervals between trials are greater than 4 seconds during sleep. Therefore, employing the jittering will not result in the blending of information from other trials. In our implementation, when fetching a sample, we randomly choose a shift step between 0 and 0.2 seconds, then shift the sample either to the left or right.

\section{Analysis of Slow Oscillation Event}
\label{sec:supp-so-analysis}
\subsection{Stages of Slow Oscillation Event}
As illustrated in Figure \ref{fig:slow-oscillation}(a), we categorize the NREM 2/3 TMR-related samples based on the onset of audio stimuli. In our implementation, raw EEG signals are segmented into 6-second samples from -1.5s to 4.5s according to the \textbf{onset} of audio stimulus. To detect "Slow Oscillation" (SO) events, we apply the method proposed by Staresina et al. \cite{staresina2015hierarchical}. We then categorize these samples according to the detected "Slow Oscillation" event. Since multiple "Slow Oscillation" events could be detected in each sample, our classification is based on the first detected "Slow Oscillation" event. \textbf{Samples are categorized into different SO-groups according to the following operational definitions.}
\begin{itemize}
  \item \textbf{Trans-1st-Up:} The onset of the audio stimuli occurs during the transition to the "1st-Up" state of the detected "Slow Oscillation" event. The onset of audio stimuli is located within 1s before the start point of the detected "1st-Up" wave peak.
  \item \textbf{1st-Up:} The onset of the audio stimuli occurs during the "1st-Up" state of the detected "Slow Oscillation" event. Since the "1st-Up" state is not clearly defined in the literature, we define it ourselves. In the time range of one second before the start point of the conventional SO result, a waveform similar to the conventional Up-state exists. To put it more directly, there is another zero point in this time range, whereby we can determine a wave peak and take the portion above 50\% of the peak value as the "1st-Up" portion; while in the actual data analysis, we noticed that there is a second possible form of "1st-Up". That is, the aforementioned required zeros cannot be detected in the corresponding time period. Then, we use the maximum of the wave peaks existing in the corresponding time period as the reference value for division. We will present what these two different "1st-Up" states (named with peak and without "1st-Up" peak, respectively) represent in the subsequent time-frequency analysis so that the reader can understand the difference more clearly.
  \item \textbf{Trans-Down:} The onset of the audio stimuli occurs during the transition to the "Down" state of the detected "Slow Oscillation" event. This corresponds to the stage between the end of the "1st-Up" state and the beginning of the "Down" state.
  \item \textbf{Down:} The onset of the audio stimuli occurs during the "Down" state of the detected "Slow Oscillation" event. Using the traditional detection method, we can identify a broader "Down" state, but we further refine this by strictly defining the "Down" state as the portion below 50\% of the value of the trough.
  \item \textbf{Trans-2nd-Up:} The onset of the audio stimuli occurs during the transition to the "2nd-Up" state of the detected "Slow Oscillation" event. This corresponds to the stage between the end of the "Down" state and the beginning of the "2nd-Up" state.
  \item \textbf{2nd-Up:} The onset of the audio stimuli occurs during the "2nd-Up" state of the detected "Slow Oscillation" event. The "2nd-Up" state aligns with the up-state in the traditional detection method, but we define it more strictly. Only portions of the waveform that exceed 50\% of the peak value are classified as the up-state.
  \item \textbf{Post-SO:} The onset of the audio stimuli occurs after the "Slow Oscillation" event. Instead of using the traditional zero point, we define the end of the "Slow Oscillation" as the conclusion of the second up-state according to our strict criteria.
  \item \textbf{Other:} All other possibilities are accounted for in the "Other" section. This encompasses scenarios where the "Slow Oscillation" is detected in locations not specified in the above definition during the detection time period, as well as instances where no "Slow Oscillation" meeting the defined criteria occurs within the detection time period.
\end{itemize}

\subsection{Distribution of SO-Groups}
\begin{figure}[h]
  \centering
  \includegraphics[width=\linewidth]{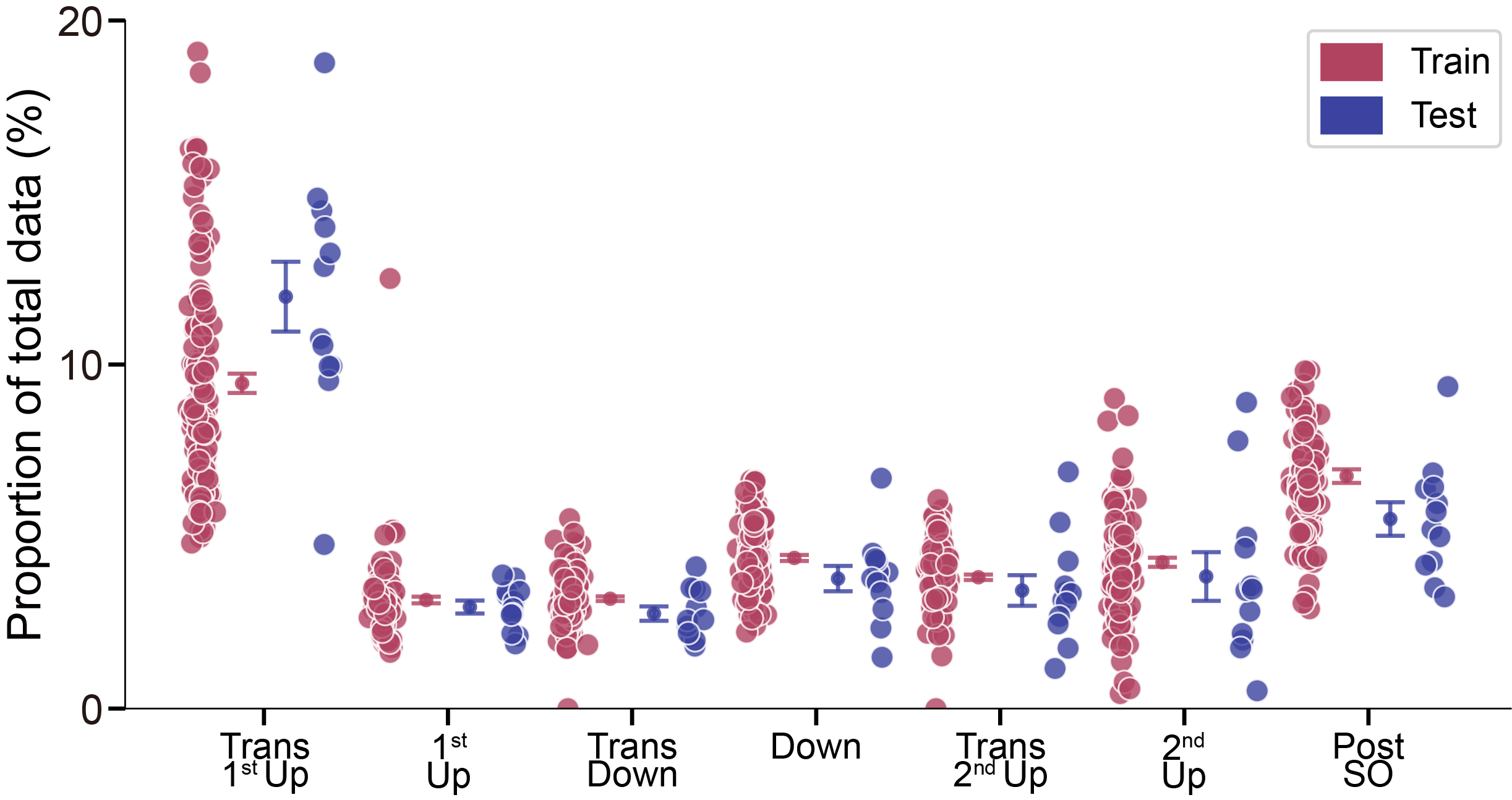}
  \caption{\textbf{The distribution of the 7 SO-groups, excluding the "Other" group, in both the training and test datasets.} The dark purple color indicates the training dataset, while the dark blue color represents the test dataset. Each dot on the graph represents a subject.}
  \label{fig:so-distribution}
\end{figure}

\subsection{Transfer Performance on different SO-Groups}

\begin{figure*}[htbp]
  \centering
  \subfigure[\label{fig:lasso-image-sleep}]{\includegraphics[width=0.9\linewidth]{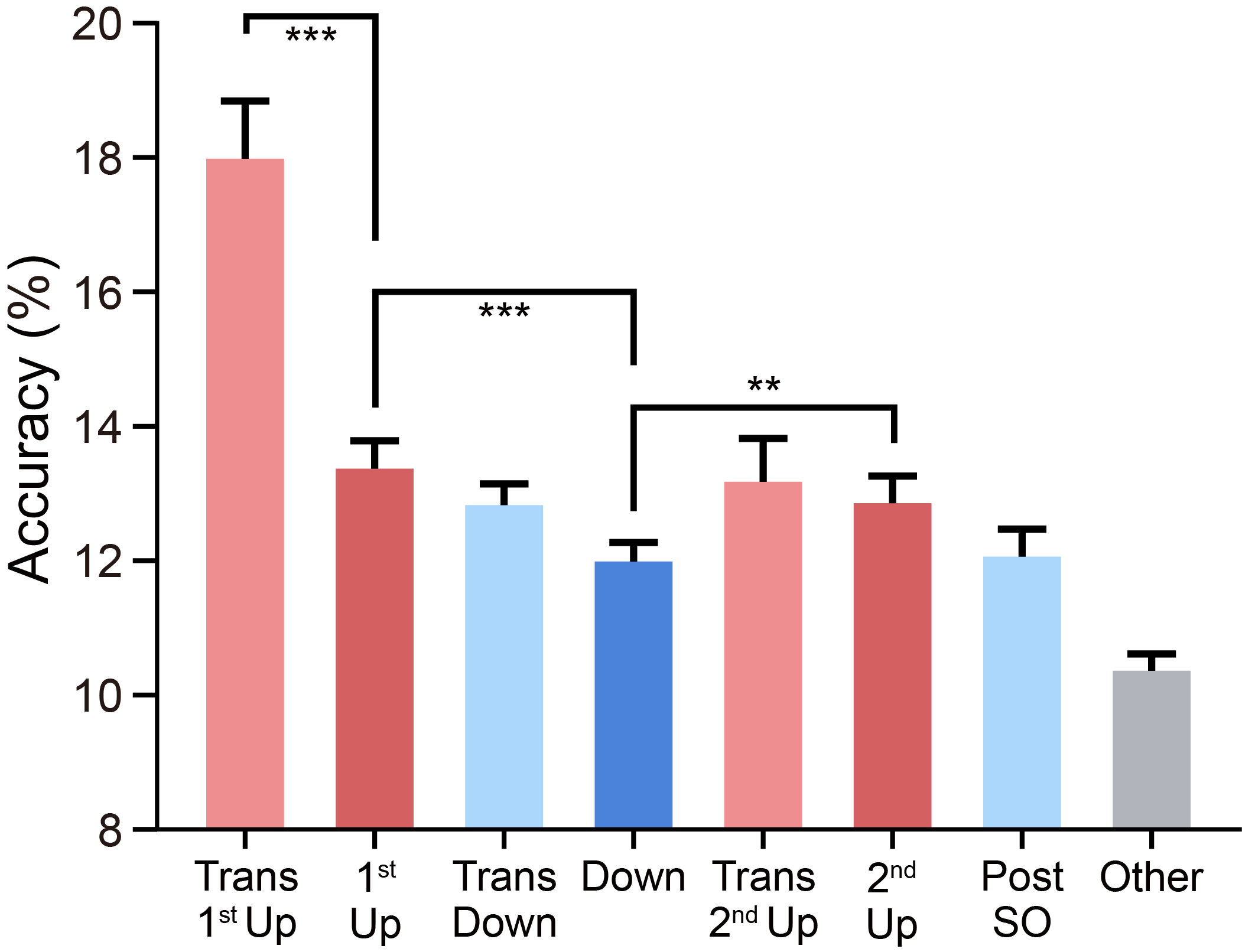}}
  \subfigure[\label{fig:lasso-audio-sleep}]{\includegraphics[width=0.9\linewidth]{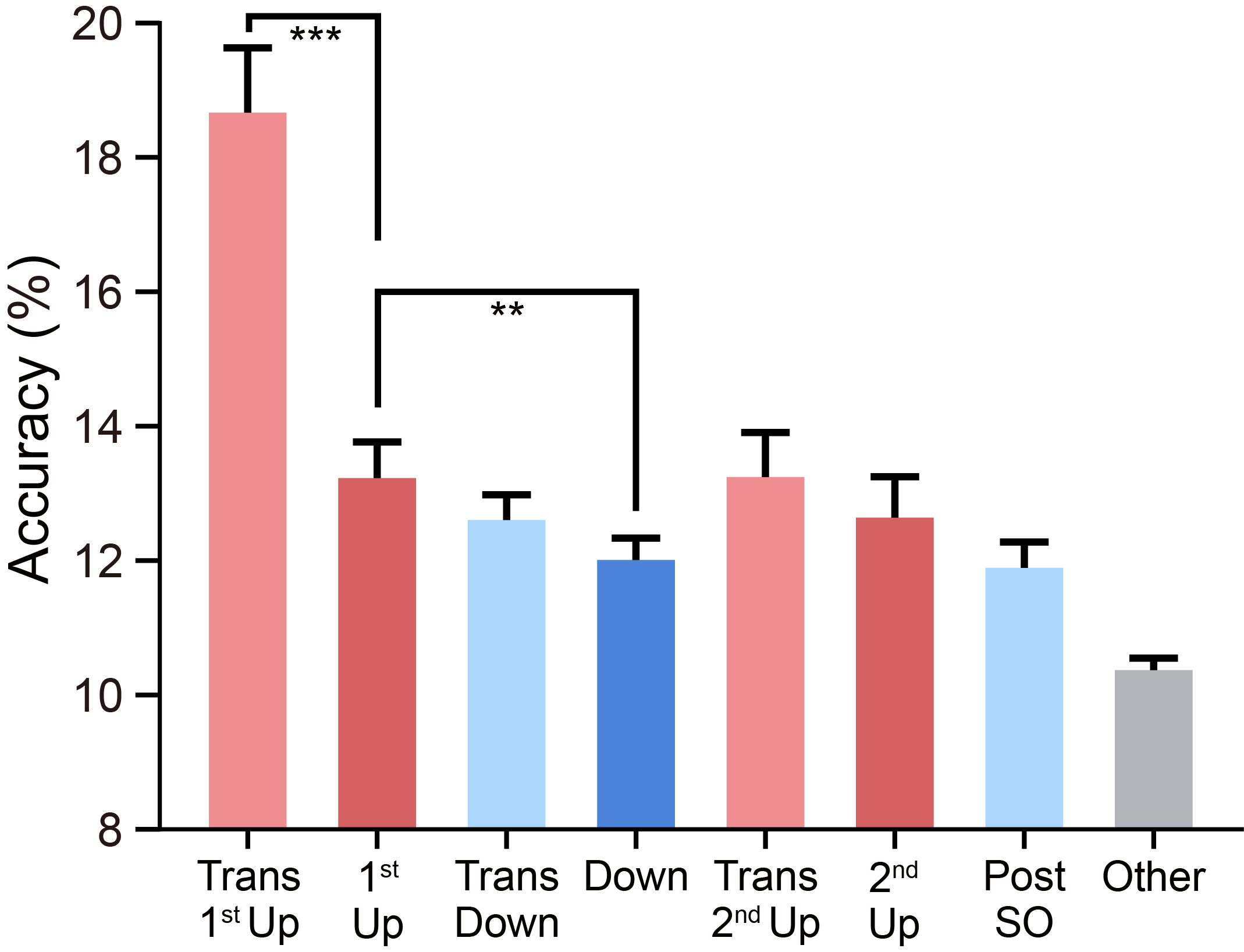}}
  \caption{\textbf{The transfer performance of Lasso-GLM \cite{horikawa2013neural} from different awake sources to NREM 2/3 sleep.} \textbf{(a).} We train Lasso-GLM on the awake image-evoked dataset $\mathcal{X}^{img}$ of one subject and evaluate it on the NREM 2/3 sleep dataset of that subject. \textbf{(b).} We train Lasso-GLM on the awake audio-evoked dataset $\mathcal{X}^{aud}$ of one subject and evaluate it on the NREM 2/3 sleep dataset of that subject.}
  \label{fig:lasso-awake-sleep}
\end{figure*}

In Figure \ref{fig:slow-oscillation}(a), we categorize samples based on the "Slow Oscillation" events and evaluate our SI-SD model on each group separately. As shown in Figure \ref{fig:slow-oscillation}(b), decoding performance varies across different groups. To determine whether these differences reflect inconsistencies in semantic information, we train a linear model, Lasso-GLM, on the awake dataset and evaluate it on different groups in NREM 2/3 sleep. The transfer performance from different awake sources (i.e., both the awake image-evoked dataset and the awake audio-evoked dataset) exhibited a similar distribution across the groups. Besides, the transfer results from the awake image-evoked dataset further prove the presence of shared semantic information between wakefulness and sleep.

\subsection{Spindle-Coupling of Slow Oscillation Event}
\begin{figure}[htbp]
  \centering
  \includegraphics[width=\linewidth]{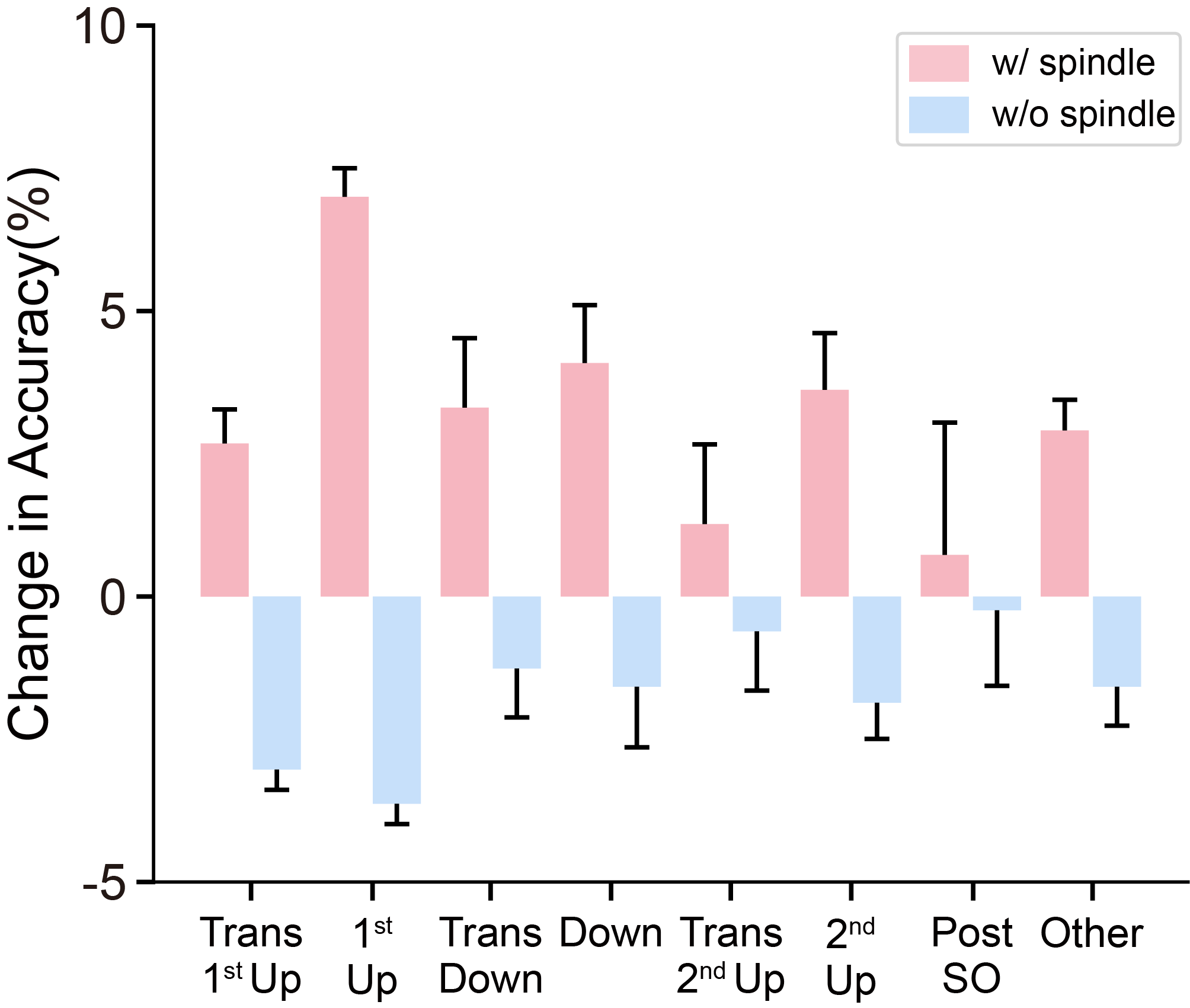}
  \caption{\textbf{The change in decoding performance for 12 subjects (w/ vs. w/o spindle coupling) in different groups.} Each SO-group is further divided into two sub-groups based on the presence or absence of spindle coupling, and the decoding accuracy is recalculated for each sub-group. Based on the decoding results from Figure \ref{fig:slow-oscillation}, the performance change of each sub-group is illustrated above, where pink represents the amount of change in the "w/ spindle" sub-group (i.e., with spindle coupling) and blue represents the amount of change in the "w/o spindle" sub-group (i.e., without spindle coupling).}
  \label{fig:spindle-performance}
\end{figure}

During our analysis of "Slow Oscillation" events, the visualization results from the temporal frequency analysis prompted us to further investigate the coupling phenomenon between "Slow Oscillation" events and spindles. Figure \ref{fig:spindle-performance} shows the changes in the performance changes of our SI-SD model on different SO-groups with or without spindle coupling. Notably, the decoding performance on the samples (w/ spindle) is higher than the samples (w/o spindle). Given the variation in decoding performance across different SO-groups, we hypothesize that the proportion of spindle coupling within these groups accounts for the differences in decoding performance and semantic information.

\begin{figure*}[htbp]
  \centering
  \subfigure[\label{fig:spindle-distribution}]{\includegraphics[width=0.9\linewidth]{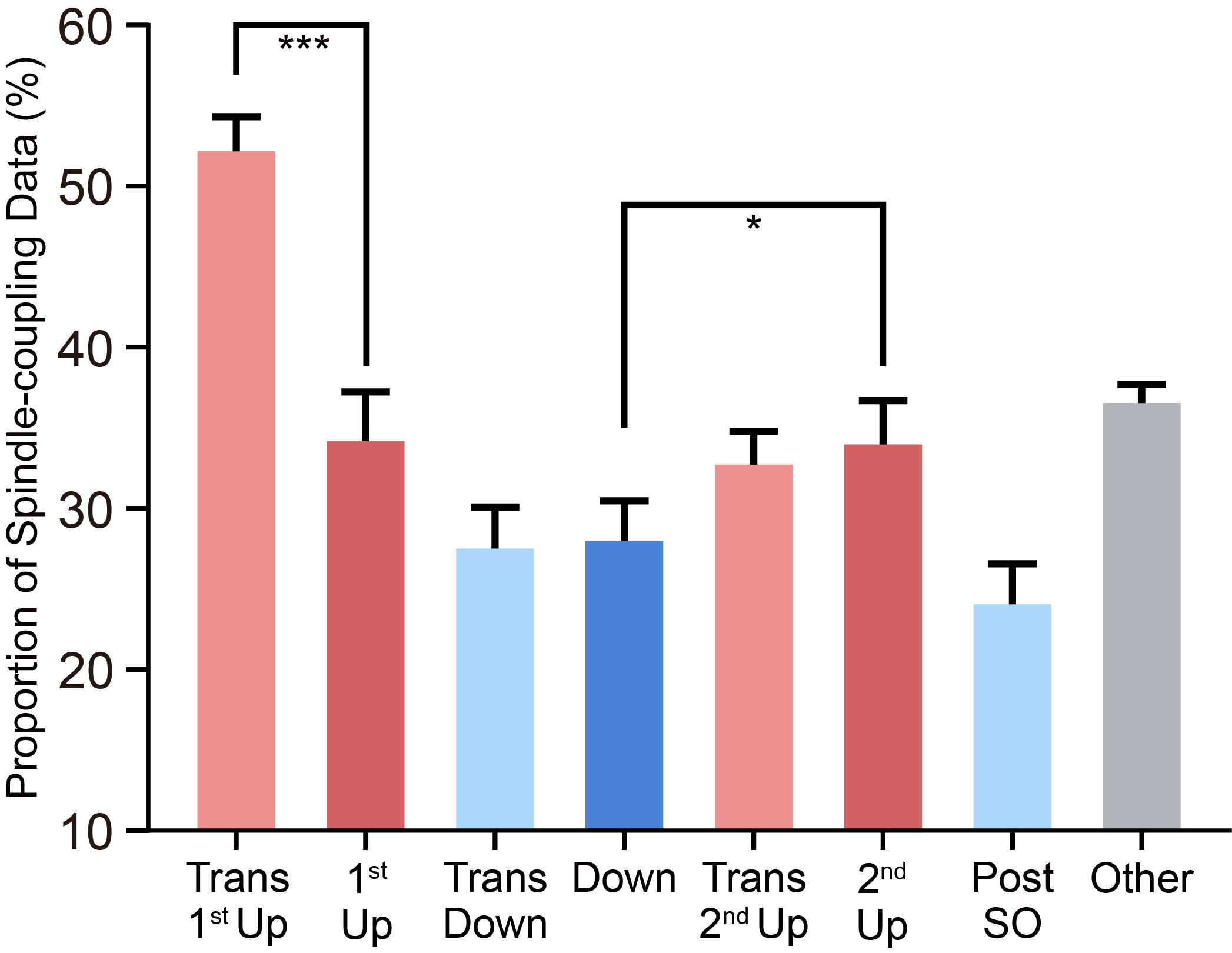}}
  \subfigure[\label{fig:so-freq-amplitude}]{\includegraphics[width=0.9\linewidth]{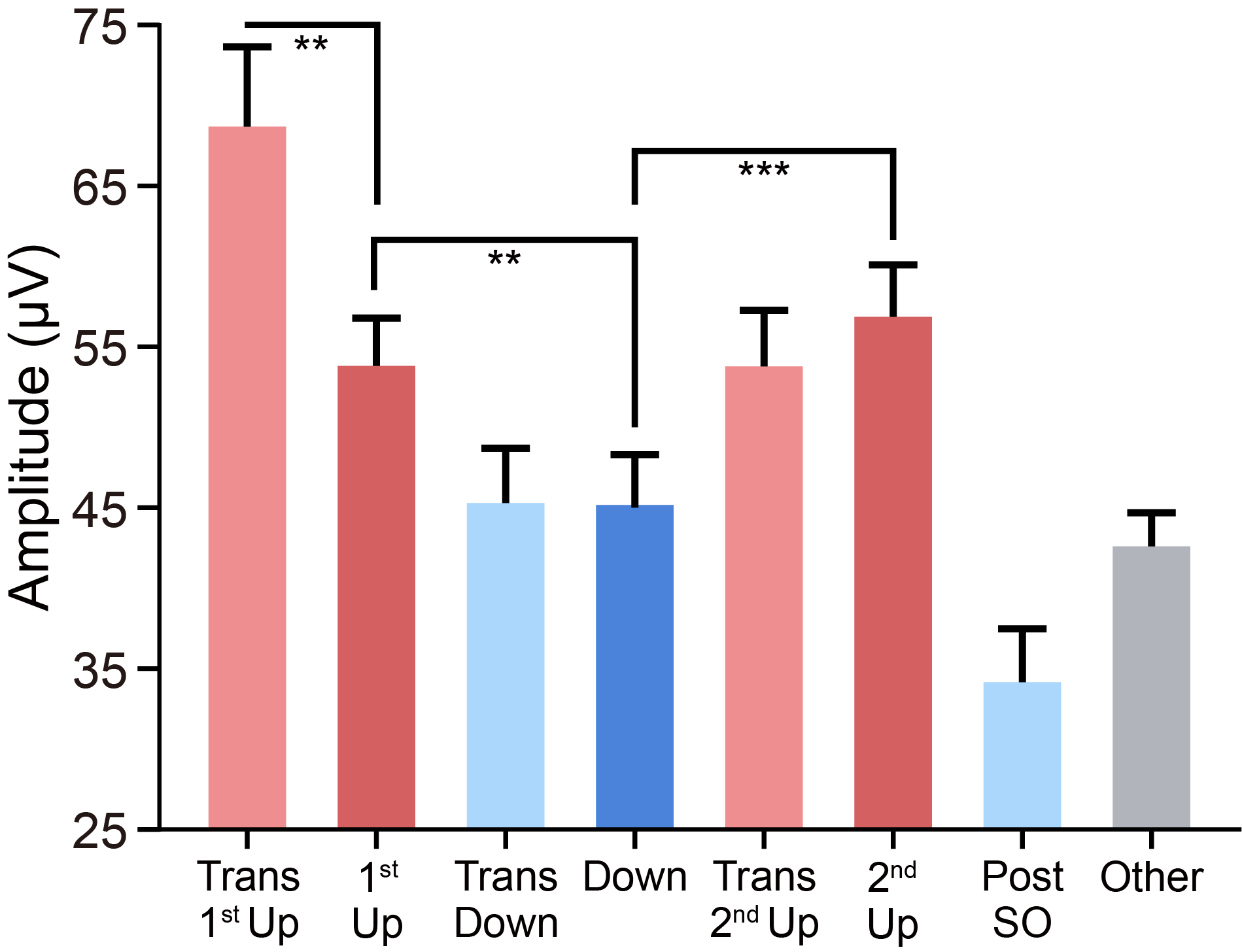}}
  \caption{\textbf{Spindle Distribution and Whole Frequency Band Amplitude Distribution.} \textbf{(a).} The distribution of SO-spindle coupling in different groups, where the differences in distribution among the groups among groups are partially consistent with previous decoding results. \textbf{(b).} The distribution of the mean amplitude in different groups. It closely mirrors existing results. The differences in distribution among the groups are statistically significant and consistent with previous decoding findings.}
  \label{fig:so-spindle-statistics}
\end{figure*}

The results in Figure \ref{fig:spindle-distribution} support our hypothesis. Additionally, the increase in coupled terms appears in the temporal frequency analysis as an enhancement of energy in the corresponding frequency band. The energy enhancement observed in the temporal frequency analysis extends well beyond the spindle frequency range (12-16 Hz). Therefore, we investigated whether the amplitude differences in the spindle bands for various data categories are consistent over a broader range or even across the full band range of the data.

Figure \ref{fig:so-freq-amplitude} validates our hypothesis, showing that full-band amplitude distributions not only resemble spindle-band amplitude distributions but also align with the statistical analysis of awake generalization results and the decoding results of the sleep pre-trained model. These findings suggest that the variability in decoding performance and the distribution of semantic information are linked to changes in signal amplitude across the full-frequency band induced by acoustic stimuli under different waveform conditions during sleep.

\subsection{Temporal-Frequency Analysis of Slow Oscillation Event}

\begin{figure*}[h]
  \centering
  \includegraphics[width=\linewidth]{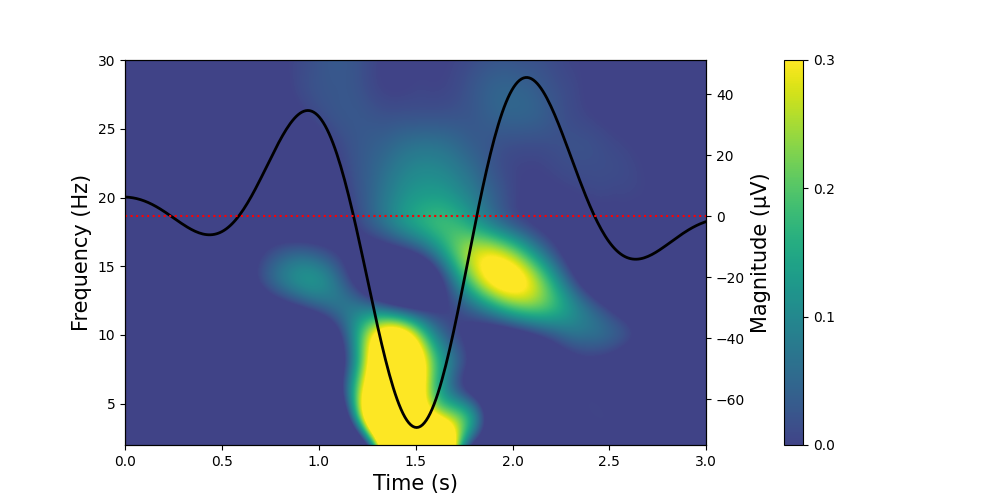}
  \includegraphics[width=\linewidth]{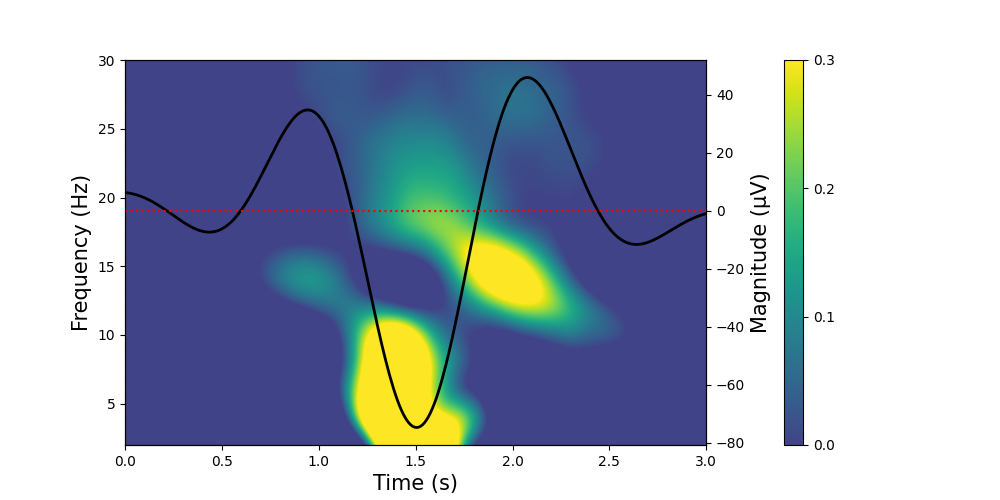}
  \includegraphics[width=\linewidth]{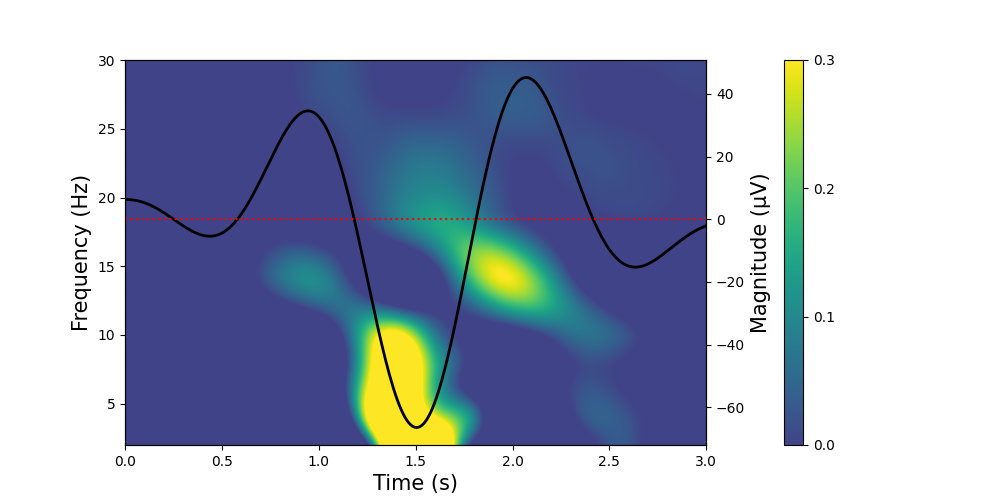}
  \caption{\textbf{The temporal frequency analysis of all "Slow Oscillation" events.} \textbf{(top).} The temporal frequency analysis of all "Slow Oscillation" events. \textbf{(middle).} The temporal frequency analysis of all "Slow Oscillation" events (w/ audio stimuli). \textbf{(bottom).} The temporal frequency analysis of all "Slow Oscillation" events (w/o audio stimuli).}
  \label{fig:tf-whole}
\end{figure*}

\begin{figure*}[h]
  \centering
  \includegraphics[width=\linewidth]{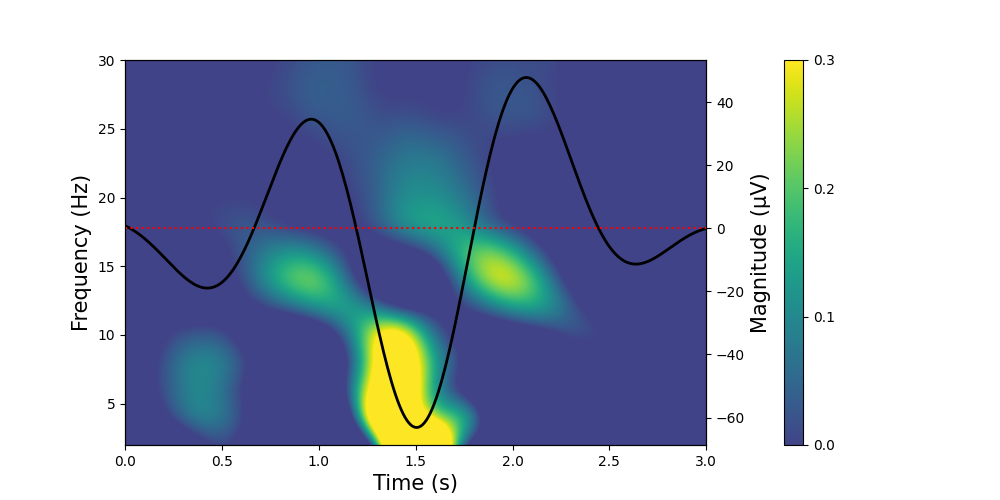}
  \includegraphics[width=\linewidth]{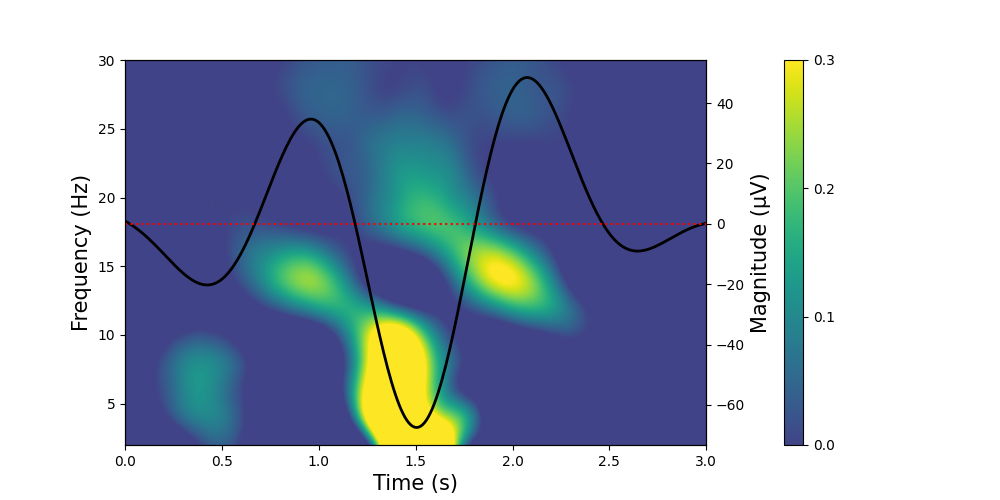}
  \includegraphics[width=\linewidth]{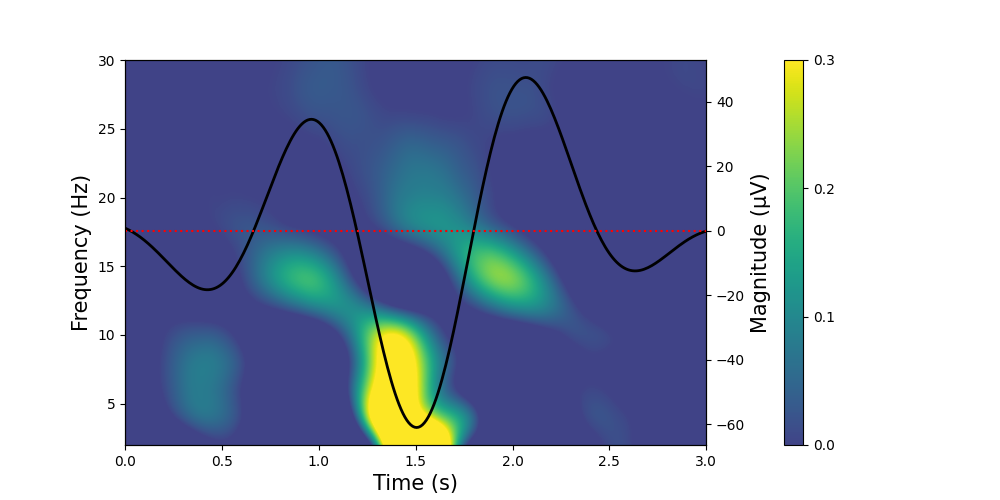}
  \caption{\textbf{The temporal frequency analysis of all "Slow Oscillation" events with "1st-Up" peak.} \textbf{(top).} The temporal frequency analysis of all "Slow Oscillation" events with "1st-Up" peak. \textbf{(middle).} The temporal frequency analysis of all "Slow Oscillation" events (w/ audio stimuli) with "1st-Up" peak. \textbf{(bottom).} The temporal frequency analysis of all "Slow Oscillation" events (w/o audio stimuli) with "1st-Up" peak.}
  \label{fig:tf-peak}
\end{figure*}

\begin{figure*}[h]
  \centering
  \includegraphics[width=\linewidth]{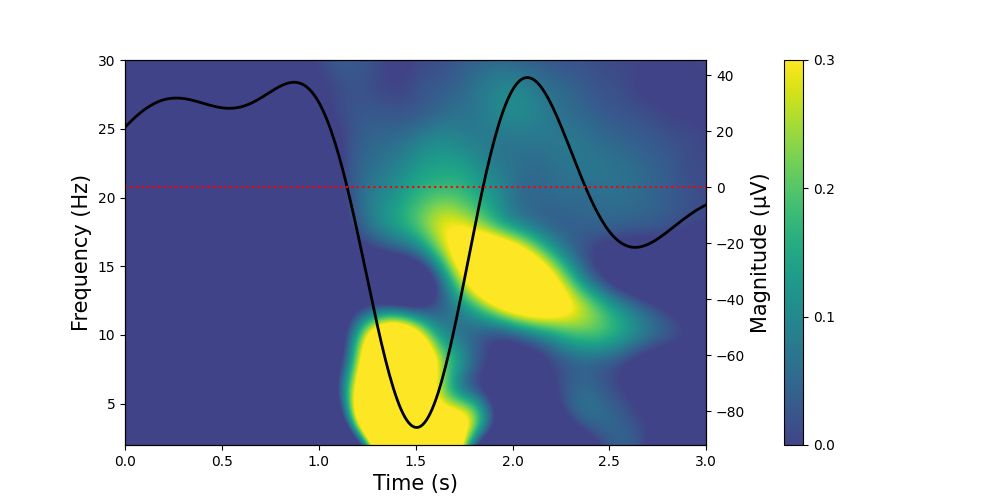}
  \includegraphics[width=\linewidth]{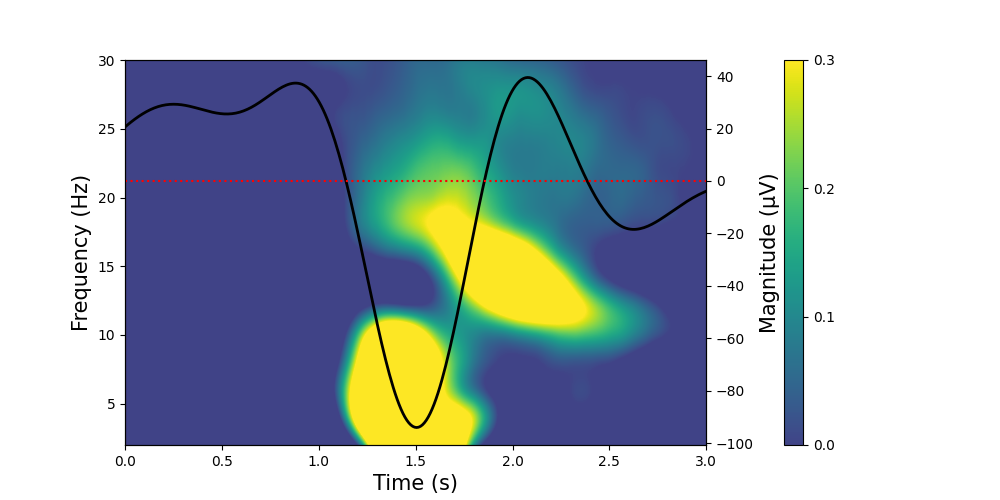}
  \includegraphics[width=\linewidth]{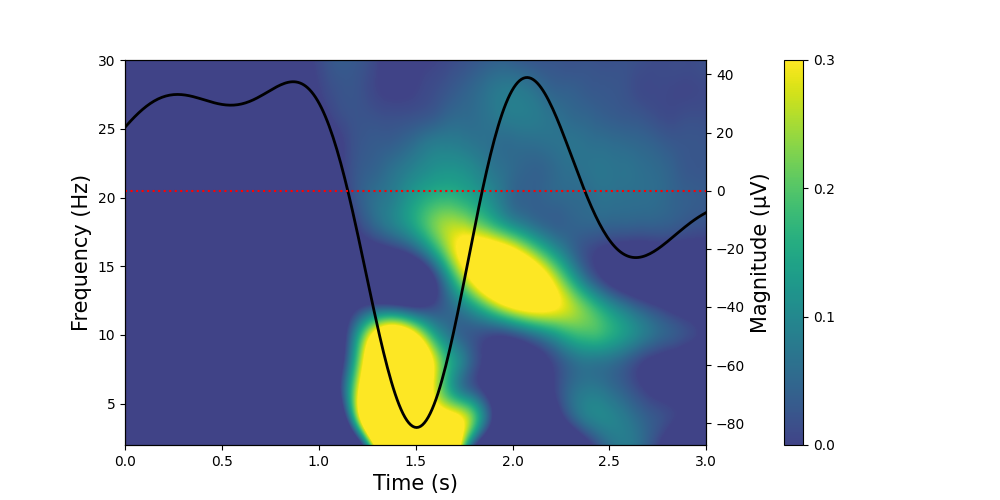}
  \caption{\textbf{The temporal frequency analysis of all "Slow Oscillation" events without "1st-Up" peak.} \textbf{(top).} The temporal frequency analysis of all "Slow Oscillation" events without "1st-Up" peak. \textbf{(middle).} The temporal frequency analysis of all "Slow Oscillation" events (w/ audio stimuli) without "1st-Up" peak. \textbf{(bottom).} The temporal frequency analysis of all "Slow Oscillation" events (w/o audio stimuli) without "1st-Up" peak.}
  \label{fig:tf-nopeak}
\end{figure*}

\begin{figure*}[h]
  \centering
  \includegraphics[width=\linewidth]{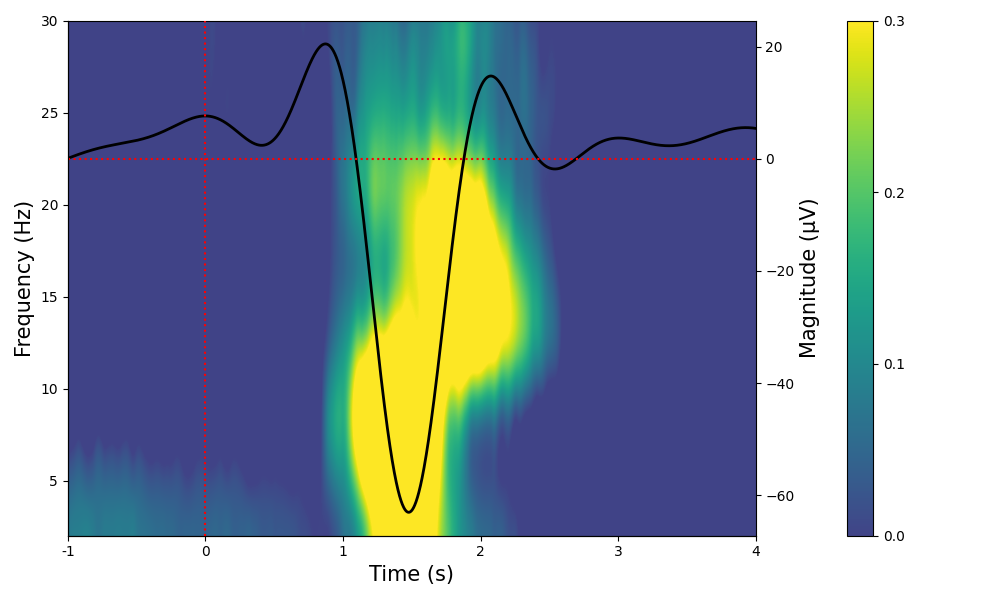}
  \caption{\textbf{The temporal frequency analysis of samples in the "Trans-1st-Up" group.}}
  \label{fig:so-trans-1st-up}
\end{figure*}

\begin{figure*}[h]
  \centering
  \includegraphics[width=\linewidth]{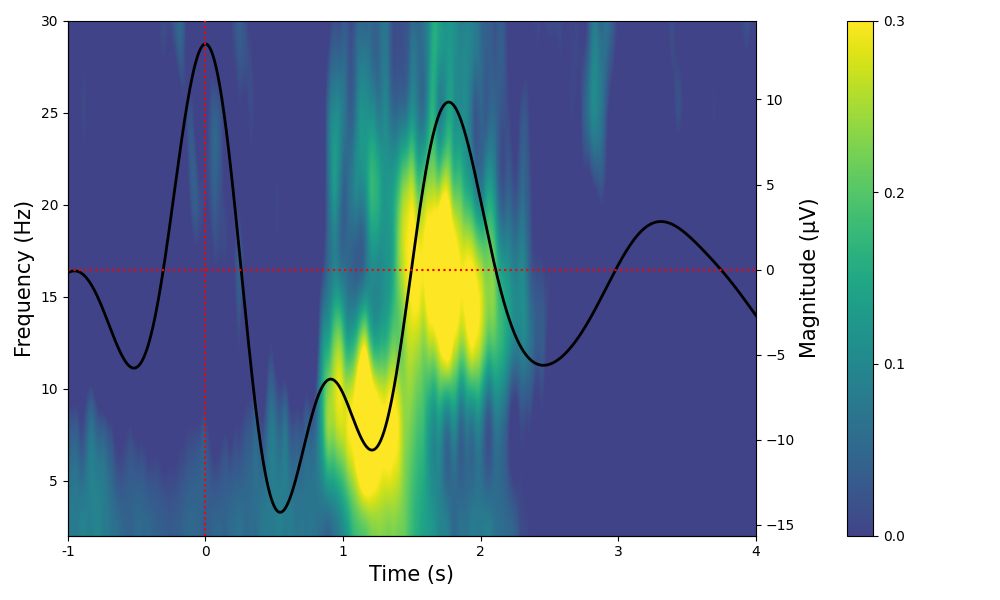}
  \caption{\textbf{The temporal frequency analysis of samples in the "1st-Up" group.}}
  \label{fig:so-1st-up}
\end{figure*}

\begin{figure*}[h]
  \centering
  \includegraphics[width=\linewidth]{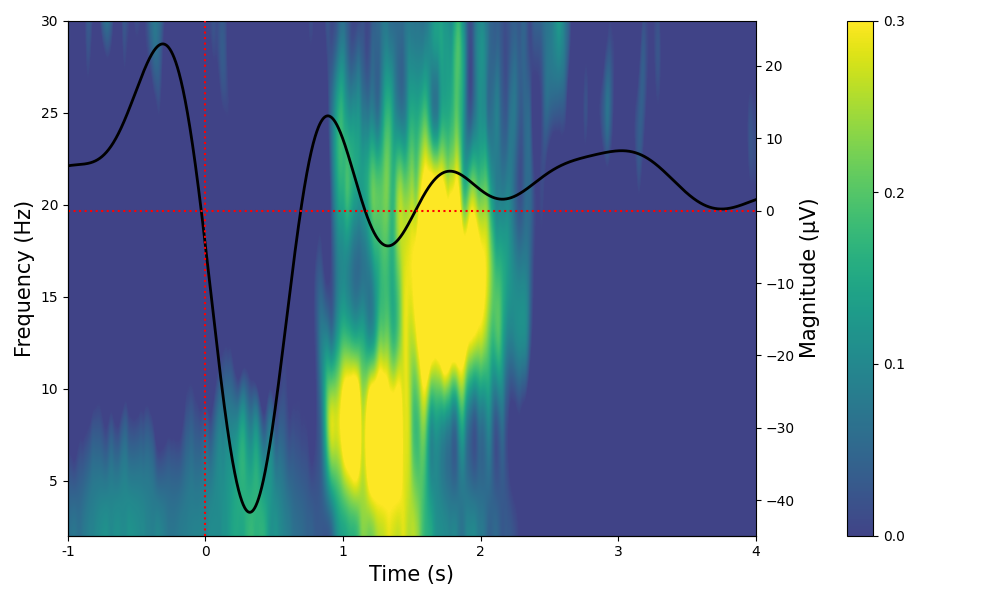}
  \caption{\textbf{The temporal frequency analysis of samples in the "Trans-Down" group.}}
  \label{fig:so-trans-down}
\end{figure*}

\begin{figure*}[h]
  \centering
  \includegraphics[width=\linewidth]{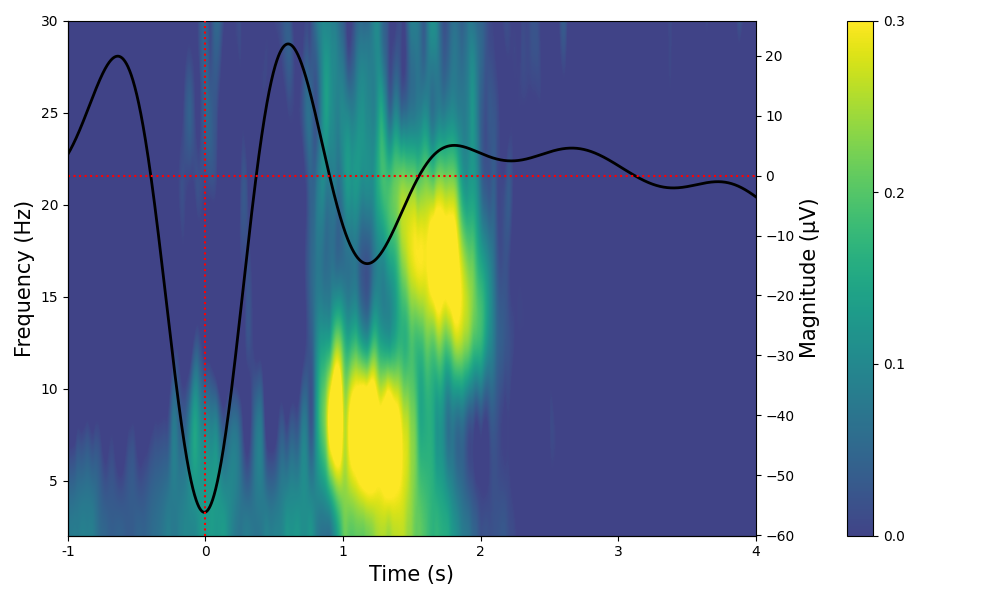}
  \caption{\textbf{The temporal frequency analysis of samples in the "Down" group.}}
  \label{fig:so-down}
\end{figure*}

\begin{figure*}[h]
  \centering
  \includegraphics[width=\linewidth]{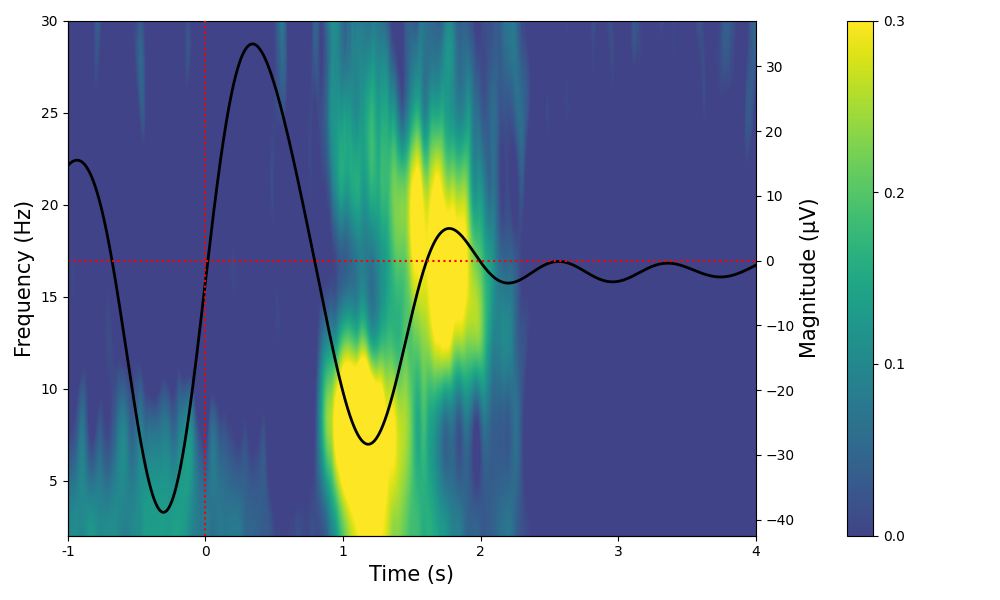}
  \caption{\textbf{The temporal frequency analysis of samples in the "Trans-2nd-Up" group.}}
  \label{fig:so-trans-2nd-up}
\end{figure*}

\begin{figure*}[h]
  \centering
  \includegraphics[width=\linewidth]{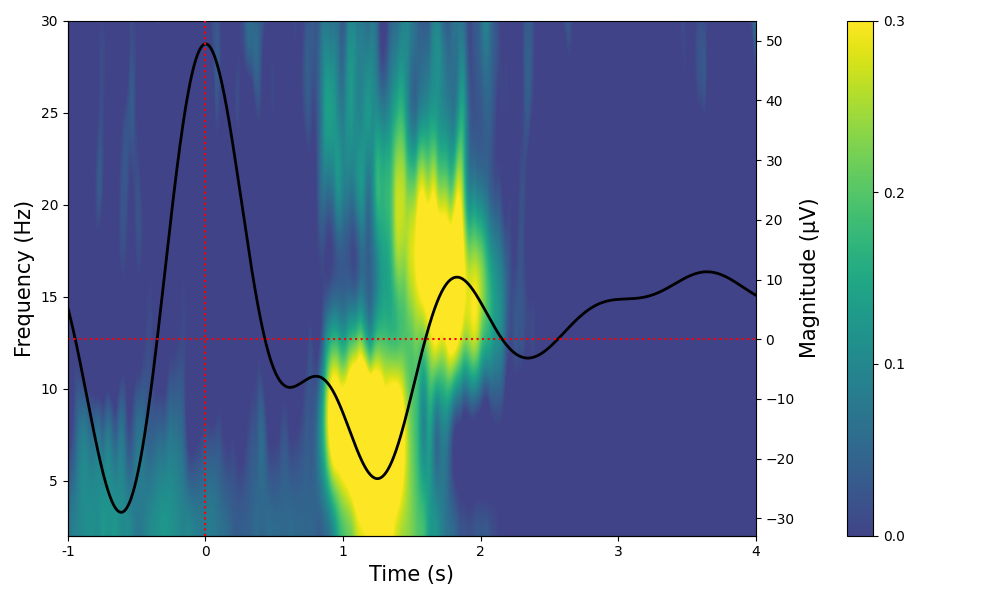}
  \caption{\textbf{The temporal frequency analysis of samples in the "2nd-Up" group.}}
  \label{fig:so-2nd-up}
\end{figure*}

\begin{figure*}[h]
  \centering
  \includegraphics[width=\linewidth]{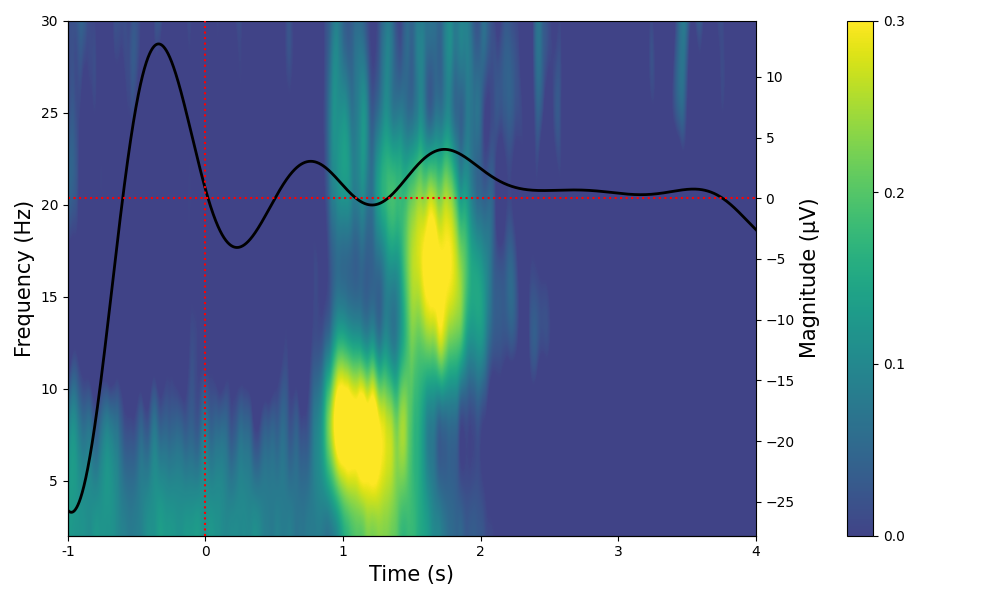}
  \caption{\textbf{The temporal frequency analysis of samples in the "Post-SO" group.}}
  \label{fig:so-post-so}
\end{figure*}

\begin{figure*}[h]
  \centering
  \includegraphics[width=\linewidth]{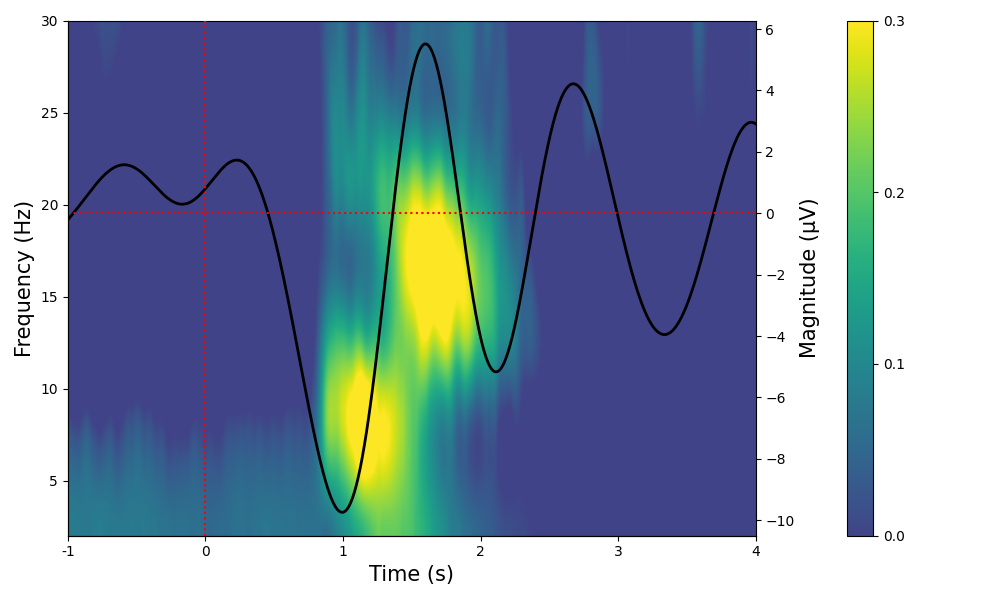}
  \caption{\textbf{The temporal frequency analysis of samples in the "Other" group.}}
  \label{fig:so-other}
\end{figure*}

\clearpage
\section{Subject-Wise Evaluation}
\label{sec:supp-subj-evaluation}
The detailed performance of different methods from each subject is provided as follows, with the best in \textbf{bold} and the second \underline{underlined}. For model comparison, we report the average and standard deviation values (within each subject) on six different random seeds to obtain comparable results.

\begin{table}[h]
  \caption{The performance of different methods from subject 01.}
  \label{table:result-subj-01}
  \centering
  \begin{tabular}{lccccc}
    \toprule
    \multirow{2}{*}{\textbf{Methods}}  & \multicolumn{2}{c}{\textbf{Zero-Shot}} & \multicolumn{2}{c}{\textbf{Fine-tune}} \\
    \cmidrule(lr){2-3}\cmidrule(lr){4-5}
    & NREM 2/3 & REM & NREM 2/3 & REM \\
    \midrule
    Lasso-GLM\cite{horikawa2013neural} & 11.90$\pm$0.34 & 12.94$\pm$0.42 & 12.52$\pm$0.32 & 16.40$\pm$0.35 \\
    \midrule
    EEG-Net\cite{lawhern2018eegnet}    & 17.00$\pm$0.87 & 17.35$\pm$0.72 & 19.55$\pm$1.08 & 28.37$\pm$0.95 \\
    EEG-Conformer\cite{song2022eeg}    & 23.24$\pm$0.92 & 12.94$\pm$0.68 & 22.75$\pm$0.92 & 23.64$\pm$1.02 \\
    \midrule
    SI-SD (single-subject)           & - & - & 21.31$\pm$0.88 &  8.95$\pm$0.34 \\
    \midrule
    SI-SD (baseline)                 & 22.33$\pm$1.22 & 18.39$\pm$0.96 & 25.49$\pm$0.81 & 31.53$\pm$0.95 \\
    +CLIP\cite{radford2021learning}    & 22.02$\pm$0.97 & 18.12$\pm$0.77 & 25.61$\pm$0.92 & 31.27$\pm$0.77 \\
    +NLA\cite{cho2023neural}           & 22.95$\pm$0.89 & 18.45$\pm$0.87 & 27.76$\pm$1.15 & 33.69$\pm$0.90 \\
    +wav2vec\cite{baevski2020wav2vec}  & \underline{24.94$\pm$0.94} & \underline{18.87$\pm$1.03} & \underline{30.76$\pm$0.79} & \underline{36.76$\pm$1.03} \\
    +Awake-Data Guide                  & \textbf{26.15$\pm$1.11} & \textbf{19.81$\pm$0.85} & \textbf{33.30$\pm$0.76} & \textbf{37.70$\pm$1.14} \\
    \bottomrule
  \end{tabular}
\end{table}

\begin{table}[h]
  \caption{The performance of different methods from subject 02.}
  \label{table:result-subj-02}
  \centering
  \begin{tabular}{lccccc}
    \toprule
    \multirow{2}{*}{\textbf{Methods}}  & \multicolumn{2}{c}{\textbf{Zero-Shot}} & \multicolumn{2}{c}{\textbf{Fine-tune}} \\
    \cmidrule(lr){2-3}\cmidrule(lr){4-5}
    & NREM 2/3 & REM & NREM 2/3 & REM \\
    \midrule
    Lasso-GLM\cite{horikawa2013neural} &  9.88$\pm$0.21 & 11.19$\pm$0.33 & 14.57$\pm$0.34 & 16.98$\pm$0.42 \\
    \midrule
    EEG-Net\cite{lawhern2018eegnet}    & 17.92$\pm$0.86 & 14.08$\pm$0.79 & 21.76$\pm$0.93 & 21.78$\pm$0.84 \\
    EEG-Conformer\cite{song2022eeg}    & 19.98$\pm$0.88 & 16.97$\pm$0.85 & 25.56$\pm$1.06 & 23.42$\pm$0.90 \\
    \midrule
    SI-SD (single-subject)           & - & - & 21.89$\pm$0.74 & 16.39$\pm$0.69 \\
    \midrule
    SI-SD (baseline)                 & 19.97$\pm$0.82 & 16.26$\pm$0.72 & 27.57$\pm$1.00 & 36.78$\pm$0.90 \\
    +CLIP\cite{radford2021learning}    & 20.19$\pm$0.92 & 16.52$\pm$0.81 & 27.51$\pm$0.87 & 36.81$\pm$0.81 \\
    +NLA\cite{cho2023neural}           & 21.02$\pm$0.80 & 19.22$\pm$0.77 & 28.01$\pm$0.92 & 37.92$\pm$0.97 \\
    +wav2vec\cite{baevski2020wav2vec}  & \underline{23.34$\pm$0.90} & \underline{19.86$\pm$0.83} & \underline{29.81$\pm$1.12} & \underline{39.46$\pm$1.21} \\
    +Awake-Data Guide                  & \textbf{24.86$\pm$0.76} & \textbf{20.94$\pm$0.79} & \textbf{30.34$\pm$1.09} & \textbf{40.47$\pm$1.34} \\
    \bottomrule
  \end{tabular}
\end{table}

\begin{table}[h]
  \caption{The performance of different methods from subject 03.}
  \label{table:result-subj-03}
  \centering
  \begin{tabular}{lccccc}
    \toprule
    \multirow{2}{*}{\textbf{Methods}}  & \multicolumn{2}{c}{\textbf{Zero-Shot}} & \multicolumn{2}{c}{\textbf{Fine-tune}} \\
    \cmidrule(lr){2-3}\cmidrule(lr){4-5}
    & NREM 2/3 & REM & NREM 2/3 & REM \\
    \midrule
    Lasso-GLM\cite{horikawa2013neural} & 11.69$\pm$0.28 & 10.13$\pm$0.30 & 12.63$\pm$0.31 & 16.39$\pm$0.41 \\
    \midrule
    EEG-Net\cite{lawhern2018eegnet}    & 16.00$\pm$0.67 & 17.26$\pm$0.72 & 22.13$\pm$0.76 & 22.12$\pm$0.73 \\
    EEG-Conformer\cite{song2022eeg}    & 18.10$\pm$0.78 & 15.93$\pm$0.71 & 24.36$\pm$0.82 & 23.98$\pm$0.88 \\
    \midrule
    SI-SD (single-subject)           & - & - & 12.84$\pm$0.64 & 11.15$\pm$0.59 \\
    \midrule
    SI-SD (baseline)                 & 19.73$\pm$0.76 & \underline{20.62$\pm$0.92} & 24.20$\pm$0.79 & 27.61$\pm$1.12 \\
    +CLIP\cite{radford2021learning}    & 19.82$\pm$0.80 & 20.02$\pm$0.74 & 24.68$\pm$0.84 & 27.19$\pm$1.04 \\
    +NLA\cite{cho2023neural}           & 19.78$\pm$0.76 & 20.34$\pm$0.79 & 25.01$\pm$0.95 & 28.57$\pm$0.94 \\
    +wav2vec\cite{baevski2020wav2vec}  & \underline{20.34$\pm$0.92} & 19.98$\pm$0.89 & \underline{27.55$\pm$1.07} & \underline{31.06$\pm$1.31} \\
    +Awake-Data Guide                  & \textbf{21.28$\pm$0.84} & \textbf{21.15$\pm$0.93} & \textbf{28.67$\pm$0.90} & \textbf{31.86$\pm$1.10} \\
    \bottomrule
  \end{tabular}
\end{table}

\begin{table}[h]
  \caption{The performance of different methods from subject 04.}
  \label{table:result-subj-04}
  \centering
  \begin{tabular}{lccccc}
    \toprule
    \multirow{2}{*}{\textbf{Methods}}  & \multicolumn{2}{c}{\textbf{Zero-Shot}} & \multicolumn{2}{c}{\textbf{Fine-tune}} \\
    \cmidrule(lr){2-3}\cmidrule(lr){4-5}
    & NREM 2/3 & REM & NREM 2/3 & REM \\
    \midrule
    Lasso-GLM\cite{horikawa2013neural} & 13.82$\pm$0.32 & 13.69$\pm$0.29 & 11.78$\pm$0.34 & 16.92$\pm$0.38 \\
    \midrule
    EEG-Net\cite{lawhern2018eegnet}    & 14.64$\pm$0.64 & 14.80$\pm$0.71 & 17.37$\pm$0.70 & 16.77$\pm$0.82 \\
    EEG-Conformer\cite{song2022eeg}    & 17.09$\pm$0.78 & 12.85$\pm$0.67 & 19.26$\pm$0.84 & 22.35$\pm$0.89 \\
    \midrule
    SI-SD (single-subject)           & - & - & 18.87$\pm$0.79 & 14.03$\pm$0.70 \\
    \midrule
    SI-SD (baseline)                 & \underline{21.12$\pm$0.97} & 15.65$\pm$0.70 & 24.09$\pm$1.02 & 22.95$\pm$0.85 \\
    +CLIP\cite{radford2021learning}    & 20.87$\pm$0.85 & 15.59$\pm$0.72 & 24.28$\pm$1.13 & 23.06$\pm$0.77 \\
    +NLA\cite{cho2023neural}           & 20.98$\pm$0.92 & 17.42$\pm$0.88 & 24.79$\pm$1.04 & 23.64$\pm$0.84 \\
    +wav2vec\cite{baevski2020wav2vec}  & 21.09$\pm$0.97 & \underline{17.60$\pm$0.92} & \underline{28.27$\pm$1.16} & \underline{23.90$\pm$0.97} \\
    +Awake-Data Guide                  & \textbf{21.18$\pm$0.76} & \textbf{20.95$\pm$0.89} & \textbf{30.98$\pm$1.21} & \textbf{24.51$\pm$1.24} \\
    \bottomrule
  \end{tabular}
\end{table}

\begin{table}[h]
  \caption{The performance of different methods from subject 05.}
  \label{table:result-subj-05}
  \centering
  \begin{tabular}{lccccc}
    \toprule
    \multirow{2}{*}{\textbf{Methods}}  & \multicolumn{2}{c}{\textbf{Zero-Shot}} & \multicolumn{2}{c}{\textbf{Fine-tune}} \\
    \cmidrule(lr){2-3}\cmidrule(lr){4-5}
    & NREM 2/3 & REM & NREM 2/3 & REM \\
    \midrule
    Lasso-GLM\cite{horikawa2013neural} & 10.59$\pm$0.31 &  9.65$\pm$0.26 & 15.26$\pm$0.38 & 16.23$\pm$0.42 \\
    \midrule
    EEG-Net\cite{lawhern2018eegnet}    & 19.07$\pm$0.65 & 13.82$\pm$0.71 & 22.43$\pm$0.84 & 13.35$\pm$0.72 \\
    EEG-Conformer\cite{song2022eeg}    & 18.37$\pm$0.78 & 14.47$\pm$0.82 & 26.68$\pm$0.94 & 24.04$\pm$0.90 \\
    \midrule
    SI-SD (single-subject)           & - & - & 20.33$\pm$0.85 & 11.13$\pm$0.72 \\
    \midrule
    SI-SD (baseline)                 & 19.94$\pm$0.82 & 13.09$\pm$0.73 & 25.52$\pm$0.87 & 22.86$\pm$0.79 \\
    +CLIP\cite{radford2021learning}    & 20.19$\pm$0.86 & 13.32$\pm$0.69 & 25.78$\pm$0.93 & 23.19$\pm$0.81 \\
    +NLA\cite{cho2023neural}           & 20.57$\pm$0.79 & 15.37$\pm$0.72 & 26.54$\pm$0.84 & 26.06$\pm$0.87 \\
    +wav2vec\cite{baevski2020wav2vec}  & \underline{21.34$\pm$0.96} & \underline{16.45$\pm$0.62} & \underline{28.38$\pm$1.03} & \underline{28.17$\pm$0.92} \\
    +Awake-Data Guide                  & \textbf{22.90$\pm$0.76} & \textbf{17.98$\pm$0.73} & \textbf{30.35$\pm$1.05} & \textbf{28.89$\pm$0.88} \\
    \bottomrule
  \end{tabular}
\end{table}

\begin{table}[h]
  \caption{The performance of different methods from subject 06.}
  \label{table:result-subj-06}
  \centering
  \begin{tabular}{lccccc}
    \toprule
    \multirow{2}{*}{\textbf{Methods}}  & \multicolumn{2}{c}{\textbf{Zero-Shot}} & \multicolumn{2}{c}{\textbf{Fine-tune}} \\
    \cmidrule(lr){2-3}\cmidrule(lr){4-5}
    & NREM 2/3 & REM & NREM 2/3 & REM \\
    \midrule
    Lasso-GLM\cite{horikawa2013neural} &  9.86$\pm$0.21 &  8.57$\pm$0.25 & 13.37$\pm$0.37 &  9.18$\pm$0.28 \\
    \midrule
    EEG-Net\cite{lawhern2018eegnet}    & 15.86$\pm$0.67 &  8.81$\pm$0.42 & 17.08$\pm$0.74 &  9.69$\pm$0.50 \\
    EEG-Conformer\cite{song2022eeg}    & 19.71$\pm$0.84 &  8.52$\pm$0.47 & 22.14$\pm$0.82 & 10.05$\pm$0.47 \\
    \midrule
    SI-SD (single-subject)           & - & - & 20.35$\pm$0.75 &  8.61$\pm$0.22 \\
    \midrule
    SI-SD (baseline)                 & 20.86$\pm$0.77 & 7.62$\pm$0.26 & 29.20$\pm$1.28 & 13.18$\pm$0.55 \\
    +CLIP\cite{radford2021learning}    & 20.57$\pm$0.82 & 7.81$\pm$0.22 & 29.03$\pm$1.09 & 13.02$\pm$0.49 \\
    +NLA\cite{cho2023neural}           & 20.98$\pm$0.89 & 8.54$\pm$0.30 & 29.75$\pm$0.95 & 13.60$\pm$0.57 \\
    +wav2vec\cite{baevski2020wav2vec}  & \underline{21.75$\pm$0.83} & \underline{9.02$\pm$0.33} & \underline{30.37$\pm$1.08} & \underline{14.54$\pm$0.57} \\
    +Awake-Data Guide                  & \textbf{23.39$\pm$0.91} & \textbf{9.52$\pm$0.31} & \textbf{30.92$\pm$0.96} & \textbf{14.91$\pm$0.64} \\
    \bottomrule
  \end{tabular}
\end{table}

\begin{table}[h]
  \caption{The performance of different methods from subject 07.}
  \label{table:result-subj-07}
  \centering
  \begin{tabular}{lccccc}
    \toprule
    \multirow{2}{*}{\textbf{Methods}}  & \multicolumn{2}{c}{\textbf{Zero-Shot}} & \multicolumn{2}{c}{\textbf{Fine-tune}} \\
    \cmidrule(lr){2-3}\cmidrule(lr){4-5}
    & NREM 2/3 & REM & NREM 2/3 & REM \\
    \midrule
    Lasso-GLM\cite{horikawa2013neural} & 11.21$\pm$0.35 & 10.14$\pm$0.30 & 13.82$\pm$0.32 & 13.56$\pm$0.29 \\
    \midrule
    EEG-Net\cite{lawhern2018eegnet}    & 17.16$\pm$0.76 & 12.70$\pm$0.57 & 18.77$\pm$0.79 & 19.15$\pm$0.81 \\
    EEG-Conformer\cite{song2022eeg}    & 19.65$\pm$0.83 & 13.24$\pm$0.59 & 19.52$\pm$0.84 & 22.83$\pm$0.90 \\
    \midrule
    SI-SD (single-subject)           & - & - & 22.31$\pm$1.03 & 10.31$\pm$0.53 \\
    \midrule
    SI-SD (baseline)                 & 19.40$\pm$0.94 & 15.31$\pm$0.70 & 25.57$\pm$1.09 & 24.35$\pm$0.76 \\
    +CLIP\cite{radford2021learning}    & 19.87$\pm$1.11 & 15.28$\pm$0.91 & 25.82$\pm$1.12 & 24.59$\pm$0.81 \\
    +NLA\cite{cho2023neural}           & 20.26$\pm$0.86 & 15.92$\pm$0.87 & 26.96$\pm$0.98 & 24.19$\pm$0.70 \\
    +wav2vec\cite{baevski2020wav2vec}  & \underline{21.86$\pm$1.01} & \underline{16.07$\pm$0.75} & \underline{28.24$\pm$1.24} & \underline{24.70$\pm$0.83} \\
    +Awake-Data Guide                  & \textbf{23.33$\pm$1.07} & \textbf{20.02$\pm$0.73} & \textbf{29.73$\pm$1.15} & \textbf{25.33$\pm$0.88} \\
    \bottomrule
  \end{tabular}
\end{table}

\begin{table}[h]
  \caption{The performance of different methods from subject 08.}
  \label{table:result-subj-08}
  \centering
  \begin{tabular}{lccccc}
    \toprule
    \multirow{2}{*}{\textbf{Methods}}  & \multicolumn{2}{c}{\textbf{Zero-Shot}} & \multicolumn{2}{c}{\textbf{Fine-tune}} \\
    \cmidrule(lr){2-3}\cmidrule(lr){4-5}
    & NREM 2/3 & REM & NREM 2/3 & REM \\
    \midrule
    Lasso-GLM\cite{horikawa2013neural} & 10.86$\pm$0.27 & 13.45$\pm$0.31 & 13.67$\pm$0.34 & 14.06$\pm$0.30 \\
    \midrule
    EEG-Net\cite{lawhern2018eegnet}    & 18.11$\pm$0.79 & 16.81$\pm$0.72 & 20.35$\pm$0.83 & 17.75$\pm$0.79 \\
    EEG-Conformer\cite{song2022eeg}    & 18.61$\pm$0.80 & 15.13$\pm$0.81 & 24.24$\pm$1.00 & 19.84$\pm$0.83 \\
    \midrule
    SI-SD (single-subject)           & - & - & 18.28$\pm$0.77 & 12.41$\pm$0.59 \\
    \midrule
    SI-SD (baseline)                 & 20.91$\pm$0.78 & 15.74$\pm$0.66 & 25.05$\pm$0.92 & 19.02$\pm$0.72 \\
    +CLIP\cite{radford2021learning}    & 20.78$\pm$0.72 & 15.82$\pm$0.58 & 24.98$\pm$0.87 & 19.19$\pm$0.69 \\
    +NLA\cite{cho2023neural}           & 21.02$\pm$0.80 & 17.36$\pm$0.71 & 25.49$\pm$0.84 & 20.96$\pm$0.93 \\
    +wav2vec\cite{baevski2020wav2vec}  & \underline{21.69$\pm$0.82} & \underline{18.49$\pm$0.77} & \underline{26.68$\pm$1.03} & \underline{21.33$\pm$0.87} \\
    +Awake-Data Guide                  & \textbf{22.53$\pm$0.94} & \textbf{27.73$\pm$1.17} & \textbf{28.08$\pm$0.89} & \textbf{21.88$\pm$1.29} \\
    \bottomrule
  \end{tabular}
\end{table}

\begin{table}[h]
  \caption{The performance of different methods from subject 09.}
  \label{table:result-subj-09}
  \centering
  \begin{tabular}{lccccc}
    \toprule
    \multirow{2}{*}{\textbf{Methods}}  & \multicolumn{2}{c}{\textbf{Zero-Shot}} & \multicolumn{2}{c}{\textbf{Fine-tune}} \\
    \cmidrule(lr){2-3}\cmidrule(lr){4-5}
    & NREM 2/3 & REM & NREM 2/3 & REM \\
    \midrule
    Lasso-GLM\cite{horikawa2013neural} & 10.95$\pm$0.22 & 13.30$\pm$0.30 & 14.65$\pm$0.35 & 17.94$\pm$0.33 \\
    \midrule
    EEG-Net\cite{lawhern2018eegnet}    & 19.25$\pm$0.73 & 20.60$\pm$0.86 & 23.36$\pm$0.81 & 36.88$\pm$1.35 \\
    EEG-Conformer\cite{song2022eeg}    & 20.48$\pm$0.70 & 19.31$\pm$0.74 & 22.88$\pm$0.91 & 38.59$\pm$1.28 \\
    \midrule
    SI-SD (single-subject)           & - & - & 26.26$\pm$1.09 & 13.05$\pm$0.57 \\
    \midrule
    SI-SD (baseline)                 & 23.70$\pm$0.83 & 22.24$\pm$0.90 & 29.35$\pm$0.89 & 39.89$\pm$1.23 \\
    +CLIP\cite{radford2021learning}    & 24.02$\pm$1.04 & 22.45$\pm$0.94 & 29.65$\pm$0.91 & 39.71$\pm$1.14 \\
    +NLA\cite{cho2023neural}           & 24.98$\pm$0.99 & 24.39$\pm$1.07 & 30.89$\pm$1.17 & 40.96$\pm$1.38 \\
    +wav2vec\cite{baevski2020wav2vec}  & \underline{26.48$\pm$0.81} & \underline{25.75$\pm$0.93} & \underline{33.44$\pm$1.20} & \underline{42.58$\pm$1.35} \\
    +Awake-Data Guide                  & \textbf{29.30$\pm$1.04} & \textbf{29.61$\pm$0.97} & \textbf{36.60$\pm$1.15} & \textbf{43.67$\pm$1.46} \\
    \bottomrule
  \end{tabular}
\end{table}

\begin{table}[h]
  \caption{The performance of different methods from subject 10.}
  \label{table:result-subj-10}
  \centering
  \begin{tabular}{lccccc}
    \toprule
    \multirow{2}{*}{\textbf{Methods}}  & \multicolumn{2}{c}{\textbf{Zero-Shot}} & \multicolumn{2}{c}{\textbf{Fine-tune}} \\
    \cmidrule(lr){2-3}\cmidrule(lr){4-5}
    & NREM 2/3 & REM & NREM 2/3 & REM \\
    \midrule
    Lasso-GLM\cite{horikawa2013neural} & 11.61$\pm$0.31 & 10.92$\pm$0.25 & 14.83$\pm$0.34 & 14.44$\pm$0.37 \\
    \midrule
    EEG-Net\cite{lawhern2018eegnet}    & 18.23$\pm$0.79 & 12.66$\pm$0.54 & 20.36$\pm$0.88 & 22.12$\pm$0.80 \\
    EEG-Conformer\cite{song2022eeg}    & 20.63$\pm$0.84 & 12.66$\pm$0.59 & 25.19$\pm$0.96 & 23.98$\pm$0.97 \\
    \midrule
    SI-SD (single-subject)           & - & - & 19.67$\pm$0.85 & 13.65$\pm$0.53 \\
    \midrule
    SI-SD (baseline)                 & 19.51$\pm$0.87 & 12.98$\pm$0.48 & 24.39$\pm$0.88 & 24.05$\pm$0.83 \\
    +CLIP\cite{radford2021learning}    & 19.72$\pm$0.77 & 13.02$\pm$0.50 & 24.85$\pm$0.79 & 24.51$\pm$0.74 \\
    +NLA\cite{cho2023neural}           & 19.97$\pm$0.92 & 13.89$\pm$0.44 & 26.19$\pm$0.83 & 26.88$\pm$0.81 \\
    +wav2vec\cite{baevski2020wav2vec}  & \underline{21.89$\pm$0.89} & \underline{13.94$\pm$0.42} & \underline{28.11$\pm$1.05} & \underline{30.12$\pm$1.09} \\
    +Awake-Data Guide                  & \textbf{23.30$\pm$0.95} & \textbf{15.72$\pm$0.63} & \textbf{28.59$\pm$1.07} & \textbf{30.89$\pm$1.17} \\
    \bottomrule
  \end{tabular}
\end{table}

\begin{table}[h]
  \caption{The performance of different methods from subject 11.}
  \label{table:result-subj-11}
  \centering
  \begin{tabular}{lccccc}
    \toprule
    \multirow{2}{*}{\textbf{Methods}}  & \multicolumn{2}{c}{\textbf{Zero-Shot}} & \multicolumn{2}{c}{\textbf{Fine-tune}} \\
    \cmidrule(lr){2-3}\cmidrule(lr){4-5}
    & NREM 2/3 & REM & NREM 2/3 & REM \\
    \midrule
    Lasso-GLM\cite{horikawa2013neural} &  9.04$\pm$0.23 &  8.87$\pm$0.25 & 12.18$\pm$0.30 &  9.69$\pm$0.27 \\
    \midrule
    EEG-Net\cite{lawhern2018eegnet}    & 17.09$\pm$0.74 & 14.41$\pm$0.69 & 18.41$\pm$0.88 & 21.33$\pm$0.96 \\
    EEG-Conformer\cite{song2022eeg}    & 18.75$\pm$0.80 & 16.08$\pm$0.73 & 23.26$\pm$1.07 & 22.84$\pm$0.94 \\
    \midrule
    SI-SD (single-subject)           & - & - & 19.59$\pm$0.85 & 13.74$\pm$0.62 \\
    \midrule
    SI-SD (baseline)                 & 21.12$\pm$1.00 & 18.25$\pm$0.75 & 26.53$\pm$0.91 & 26.84$\pm$0.75 \\
    +CLIP\cite{radford2021learning}    & 21.02$\pm$0.92 & 18.22$\pm$0.68 & 26.41$\pm$0.89 & 26.89$\pm$0.81 \\
    +NLA\cite{cho2023neural}           & 21.56$\pm$0.89 & 18.84$\pm$0.77 & 26.88$\pm$0.87 & 27.21$\pm$0.95 \\
    +wav2vec\cite{baevski2020wav2vec}  & \underline{22.49$\pm$1.08} & \underline{18.96$\pm$0.81} & \underline{27.52$\pm$1.12} & 28.16$\pm$1.06 \\
    +Awake-Data Guide                  & \textbf{23.76$\pm$0.98} & \textbf{19.58$\pm$0.94} & \textbf{28.70$\pm$1.19} & \textbf{29.91$\pm$1.10} \\
    \bottomrule
  \end{tabular}
\end{table}

\begin{table}[h]
  \caption{The performance of different methods from subject 12.}
  \label{table:result-subj-12}
  \centering
  \begin{tabular}{lccccc}
    \toprule
    \multirow{2}{*}{\textbf{Methods}}  & \multicolumn{2}{c}{\textbf{Zero-Shot}} & \multicolumn{2}{c}{\textbf{Fine-tune}} \\
    \cmidrule(lr){2-3}\cmidrule(lr){4-5}
    & NREM 2/3 & REM & NREM 2/3 & REM \\
    \midrule
    Lasso-GLM\cite{horikawa2013neural} & 12.99$\pm$0.34 & 13.96$\pm$0.33 & 15.62$\pm$0.37 & 16.54$\pm$0.32 \\
    \midrule
    EEG-Net\cite{lawhern2018eegnet}    & 17.29$\pm$0.76 & 16.79$\pm$0.59 & 15.76$\pm$0.58 & 18.34$\pm$0.65 \\
    EEG-Conformer\cite{song2022eeg}    & 22.20$\pm$0.97 & 17.47$\pm$0.76 & 21.98$\pm$0.98 & \underline{25.03$\pm$1.18} \\
    \midrule
    SI-SD (single-subject)           & - & - & 19.67$\pm$0.79 & 18.32$\pm$0.72 \\
    \midrule
    SI-SD (baseline)                 & 23.48$\pm$0.89 & 18.01$\pm$0.77 & 26.42$\pm$1.02 & 28.45$\pm$0.89 \\
    +CLIP\cite{radford2021learning}    & 23.15$\pm$0.74 & 17.95$\pm$0.79 & 26.09$\pm$0.99 & 28.31$\pm$0.77 \\
    +NLA\cite{cho2023neural}           & 24.42$\pm$0.97 & 19.39$\pm$0.83 & 26.79$\pm$0.82 & 28.96$\pm$1.03 \\
    +wav2vec\cite{baevski2020wav2vec}  & \underline{25.83$\pm$1.04} & \underline{19.68$\pm$0.90} & \underline{27.36$\pm$1.19} & \underline{29.45$\pm$1.17} \\
    +Awake-Data Guide                  & \textbf{27.49$\pm$1.24} & \textbf{20.21$\pm$0.93} & \textbf{27.56$\pm$1.11} & \textbf{30.21$\pm$1.21} \\
    \bottomrule
  \end{tabular}
\end{table}

\clearpage
\section*{NeurIPS Paper Checklist}
\begin{enumerate}
\item {\bf Claims}
  \item[] Question: Do the main claims made in the abstract and introduction accurately reflect the paper's contributions and scope?
  \item[] Answer: \answerYes{} 
  \item[] Justification: Three of the four contributions mentioned at the end of the "Introduction" section are explicitly included. The contribution related to neuroscience-inspired analysis is simplified as "inspired by neuroscience findings" at the end of the "Abstraction" section.

\item {\bf Limitations}
  \item[] Question: Does the paper discuss the limitations of the work performed by the authors?
  \item[] Answer: \answerYes{} 
  \item[] Justification: We provide a separate "Limitations" section; see Section \ref{sec:limitations} for more details.

\item {\bf Theory Assumptions and Proofs}
  \item[] Question: For each theoretical result, does the paper provide the full set of assumptions and a complete (and correct) proof?
  \item[] Answer: \answerNA{} 
  \item[] Justification: The paper does not include theoretical results.

\item {\bf Experimental Result Reproducibility}
  \item[] Question: Does the paper fully disclose all the information needed to reproduce the main experimental results of the paper to the extent that it affects the main claims and/or conclusions of the paper (regardless of whether the code and data are provided or not)?
  \item[] Answer: \answerYes{} 
  \item[] Justification: We provide detailed information related to our model and baselines in Appendix \ref{sec:supp-model-details} and Appendix \ref{sec:supp-baseline-details}, respectively. Besides, we provide code and demo dataset in Section \ref{sec:reprod-state}.

\item {\bf Open access to data and code}
  \item[] Question: Does the paper provide open access to the data and code, with sufficient instructions to faithfully reproduce the main experimental results, as described in supplemental material?
  \item[] Answer: \answerYes{} 
  \item[] Justification: We provide code and demo dataset in Section \ref{sec:reprod-state}. Due to the lack of open-source sEEG datasets related to language, we collected a well-annotated Chinese word-reading sEEG dataset, and evaluated our model on this dataset. However, respecting the efforts of the data collectors, we only provide the dataset of some subjects to reproduce the experimental results in the main text. The whole dataset will be publicly available to ensure the reproducibility of this work.

\item {\bf Experimental Setting/Details}
  \item[] Question: Does the paper specify all the training and test details (e.g., data splits, hyperparameters, how they were chosen, type of optimizer, etc.) necessary to understand the results?
  \item[] Answer: \answerYes{} 
  \item[] Justification: We provide detailed information related to our model and baselines in Appendix \ref{sec:supp-model-details} and Appendix \ref{sec:supp-baseline-details}, respectively.

\item {\bf Experiment Statistical Significance}
  \item[] Question: Does the paper report error bars suitably and correctly defined or other appropriate information about the statistical significance of the experiments?
  \item[] Answer: \answerYes{} 
  \item[] Justification: For the main results, we report the average and standard error values (of all subjects) on six random seeds. For detailed subject-wise evaluation, we report the average and standard deviation values (of each subject) on six random seeds.

\item {\bf Experiments Compute Resources}
  \item[] Question: For each experiment, does the paper provide sufficient information on the computer resources (type of compute workers, memory, time of execution) needed to reproduce the experiments?
  \item[] Answer: \answerYes{} 
  \item[] Justification: Detailed information related to the training process is provided in Section \ref{sec:imple-details}
    
\item {\bf Code Of Ethics}
  \item[] Question: Does the research conducted in the paper conform, in every respect, with the NeurIPS Code of Ethics \url{https://neurips.cc/public/EthicsGuidelines}?
  \item[] Answer: \answerYes{} 
  \item[] Justification: The research conducted in the paper conforms, in every respect, with the NeurIPS Code of Ethics \url{https://neurips.cc/public/EthicsGuidelines}.

\item {\bf Broader Impacts}
  \item[] Question: Does the paper discuss both potential positive societal impacts and negative societal impacts of the work performed?
  \item[] Answer: \answerNo{} 
  \item[] Justification: This work aims to explore the feasibility of EEG to decode sleep, which mainly has positive impacts on society.
    
\item {\bf Safeguards}
  \item[] Question: Does the paper describe safeguards that have been put in place for responsible release of data or models that have a high risk for misuse (e.g., pretrained language models, image generators, or scraped datasets)?
  \item[] Answer: \answerNo{} 
  \item[] Justification: The paper poses no such risks.

\item {\bf Licenses for existing assets}
  \item[] Question: Are the creators or original owners of assets (e.g., code, data, models), used in the paper, properly credited and are the license and terms of use explicitly mentioned and properly respected?
  \item[] Answer: \answerYes{} 
  \item[] Justification: The creators or original owners of assets (e.g., code, data, models), used in the paper, are properly credited and are the license and terms of use explicitly mentioned and properly respected.

\item {\bf New Assets}
  \item[] Question: Are new assets introduced in the paper well documented and is the documentation provided alongside the assets?
  \item[] Answer: \answerNA{} 
  \item[] Justification: The paper does not release new assets.

\item {\bf Crowdsourcing and Research with Human Subjects}
  \item[] Question: For crowdsourcing experiments and research with human subjects, does the paper include the full text of instructions given to participants and screenshots, if applicable, as well as details about compensation (if any)? 
  \item[] Answer: \answerYes{} 
  \item[] Justification: The ethics statements are provided in Section \ref{sec:ethics-state}.

\item {\bf Institutional Review Board (IRB) Approvals or Equivalent for Research with Human Subjects}
  \item[] Question: Does the paper describe potential risks incurred by study participants, whether such risks were disclosed to the subjects, and whether Institutional Review Board (IRB) approvals (or an equivalent approval/review based on the requirements of your country or institution) were obtained?
  \item[] Answer: \answerYes{} 
  \item[] Justification: The ethics statements are provided in Section \ref{sec:ethics-state}.
\end{enumerate}

\end{document}